\documentclass[10pt,journal,compsoc]{IEEEtran}

\usepackage{colortbl}
\usepackage{graphicx}
\usepackage{subcaption}
\usepackage{xspace}
\usepackage{multirow}
\usepackage{makecell}
\usepackage{algorithm}
\usepackage{amsmath}
\usepackage{enumitem}

\makeatletter
\DeclareRobustCommand\onedot{\futurelet\@let@token\@onedot}
\def\@onedot{\ifx\@let@token.\else.\null\fi\xspace}

\makeatother

\definecolor{yellow}{rgb}{1, 1, 0.7}
\definecolor{orange}{rgb}{1, 0.85, 0.7}
\definecolor{tablered}{rgb}{1, 0.7, 0.7}
\definecolor{red}{rgb}{1, 0, 0}

\definecolor{wincolor}{rgb}{0.85, 0.0, 0.0}

\definecolor{darkyellow}{rgb}{0.8, 0.8, 0.5}
\definecolor{darkred}{rgb}{0.7, 0.3, 0.3}
\definecolor{darkgreen}{rgb}{0.3, 0.7, 0.3}
\definecolor{green}{rgb}{0, 1.0, 0}
\definecolor{pink}{rgb}{1, 0.4, 0.7}

\usepackage{graphicx}
\usepackage{subcaption}
\usepackage{xcolor}
\usepackage{amsmath}
\usepackage{booktabs}
\usepackage{ulem}
\usepackage{tablefootnote}
\usepackage{amsfonts,amssymb}
\usepackage{comment}
\usepackage{float}
\usepackage{hyperref}
\usepackage{pdfpages}
\renewcommand{\arraystretch}{1.4}

\ifCLASSOPTIONcompsoc
  \usepackage[nocompress]{cite}
\else
  \usepackage{cite}
\fi

\ifCLASSINFOpdf
\else
\fi

\begin{document}
\title{CAD-MLLM: Unifying Multimodality-Conditioned CAD Generation With MLLM}
 
\author{Jingwei~Xu*,
        Chenyu~Wang*,
        Zibo~Zhao,
        Wen~Liu,
        Yi~Ma,
        Shenghua~Gao%
\IEEEcompsocitemizethanks{
\IEEEcompsocthanksitem Jingwei Xu and Chenyu Wang contributed equally to this work;
\IEEEcompsocthanksitem  Corresponding Author: Shenghua Gao; \protect\\ 
E-mail: gaosh@hku.hk
\IEEEcompsocthanksitem Jingwei Xu and Zibo Zhao are with the School of Information Science and Technology, ShanghaiTech University, Shanghai 201210, China. Email: xujw2023@shanghaitech.edu.cn, zhaozb@shanghaitech.edu.cn
\IEEEcompsocthanksitem Chenyu Wang is with Transcengram. Email: wangchy@transcengram.com
\IEEEcompsocthanksitem Wen Liu is with DeepSeek AI. Email: liuwen@deepseek.com
\IEEEcompsocthanksitem Yi Ma, and Shenghua Gao are with the University of Hong Kong, Hong Kong SAR,
China. E-mail: mayi@hku.hk, gaosh@hku.hk
}%
}

\markboth{Journal of \LaTeX\ Class Files,~Vol.~14, No.~8, August~2015}%
{Shell \MakeLowercase{\textit{et al.}}: Bare Demo of IEEEtran.cls for Computer Society Journals}

\IEEEtitleabstractindextext{%
\begin{abstract}

This paper aims to design a unified Computer-Aided Design (CAD) generation system that can easily generate CAD models based on the user's inputs in the form of textual description, images, point clouds, or even a combination of them. Towards this goal, we introduce the CAD-MLLM, the first system capable of generating parametric CAD models conditioned on the multimodal input. Specifically, within the CAD-MLLM framework, we leverage the command sequences of CAD models and then employ advanced large language models (LLMs) to align the feature space across these diverse multi-modalities data and CAD models' vectorized representations. To facilitate the model training, we design a comprehensive data construction and annotation pipeline that equips each CAD model with corresponding multimodal data. Our resulting dataset, named Omni-CAD, is the first multimodal CAD dataset that contains textual description, multi-view images, points, and command sequence for each CAD model. It contains approximately 450K instances and their CAD construction sequences. To thoroughly evaluate the quality of our generated CAD models, we go beyond current evaluation metrics that focus on reconstruction quality by introducing additional metrics that assess topology quality and surface enclosure extent. Extensive experimental results demonstrate that CAD-MLLM significantly outperforms existing conditional generative methods and remains highly robust to noises and missing points. The project page and
more visualizations can be found at: \href{https://cad-mllm.github.io/}{https://cad-mllm.github.io/}

\end{abstract}

\begin{IEEEkeywords}
Computer-Aided Design Models, Multimodal Large Language Models, Multimodality Data
\end{IEEEkeywords}}

\maketitle

\IEEEdisplaynontitleabstractindextext

\IEEEpeerreviewmaketitle

\IEEEraisesectionheading{\section{Introduction}\label{sec:introduction}}

Computer-Aided Design (CAD) is the use of computers to aid the creation, modification, and optimization of objects. It plays a pivotal role in industrial design and manufacturing and has been widely used for architectural design, shipbuilding, automobile, and aerospace industries, etc. Classical CAD workflow usually involves the design of 2D sketches (e.g., circles, lines, splines) and 3D operations (e.g., extrusion, loft, fillet) of these 2D elements, which is a sequence of actions with fixed types but various parameters, which can be explicitly and precisely represented by text. Then, the final CAD models are shaved as boundary representations (B-Rep), which facilitates the control of the design history and modification of the models. However, current CAD software requires experts to design and modify the model, while the CAD needs to be frequently updated by communicating with the users. It is desirable to develop a toolbox with which the expert, or even the non-expert, can easily design the CAD models by using simple instructions and illustrations to make the ideas in their mind easily come true.

With the advancement of generative models, recent approaches have explored CAD generation, of which, DeepCAD~\cite{DeepCAD} is a very representative one. DeepCAD leverages an autoencoder to generate CAD models from command sequence representation, but it operates exclusively within a latent space and is not initially designed for conditional generation, which could not meet the users' needs for interactive design, where the users' input can be images, textual descriptions, or point clouds. To tackle this issue, Img2CAD\cite{you2024img2cad} and GenCAD~\cite{alam2024gencad} have been proposed to generate a CAD model based on the input images. Text2CAD~\cite{khan2024text2cadgeneratingsequentialcad}, Query2CAD~\cite{badagabettu2024query2cadgeneratingcadmodels} have been proposed to generate a CAD model based on the text. Point2cyl~\cite{uy2022point2cyl}, TransCAD~\cite{dupont2024transcadhierarchicaltransformercad} have been proposed to generate a CAD model based on the point cloud. However, all these methods propose different methods for conditions of different modalities. It is desirable to design a unified framework to tackle the CAD generation task with different input conditions or even multiple conditions. 

\begin{table*}[t]
\centering
\renewcommand\arraystretch{1.0}
\resizebox{2\columnwidth}{!}
{
\begin{tabular}{cccccc}
\toprule
Dataset                  & Publication    & CAD Model Representation$\ddagger$ & CAD Model  Size  & Input Condition \\ 
\midrule
ABC~\cite{ABC} & CVPR 2019 &	B-rep	& 1,000,000+ & Uncondition\\
CC3D-Ops~\cite{dupont2022cadops} & 3DV 2022 & B-rep	& 37,000+ & Uncondition\\
CADParser~\cite{zhou2023cadparser} &	IJCAI 2023	& Command Sequence & 40,000+ & Uncondition \\
DeepCAD~\cite{DeepCAD} & ICCV 2021 & Command Sequence & 179,133 & Uncondition\\
Fusion360~\cite{fusion360recon} & TOG 2021 & Command Sequence & 8,625 & Uncondition\\
SketchGraphs~\cite{SketchGraphs}$\dagger$  & Arxiv, 2020.7 & Command Sequence & 15,000,000+ & Uncondition\\
\hline
Free2CAD~\cite{li2022free2cad} & SIGGRAGH 2022	& Command Sequence	& 210,000+	& User Drawing \\
Img2CAD~\cite{you2024img2cad}  & Arxiv, 2024.7 & Command Sequence & 4,574 & Single Image\\
OpenECAD~\cite{Yuan_2024}  & Arxiv, 2024.6 & Command Sequence & 200,000+ & Single Image\\
ABC-mono~\cite{chen2024img2cad}  & Arxiv, 2024.10 & Command Sequence & 208,853 & Single Image\\
\hline
Query2CAD~\cite{badagabettu2024query2cadgeneratingcadmodels}  & Arxiv, 2024.5 & Python Macro & 57 & Text\\
Text2CAD~\cite{{khan2024text2cadgeneratingsequentialcad}}  & NeurIPS 2024 & Command Sequence & 158,000+ & Text\\
\hline
Omni-CAD(Ours)  &  & Command Sequence & 453,220 & Multi-view Images/Text/Point \\
\bottomrule
\end{tabular}
}

\caption{Comparison of previous datasets and our proposed dataset. Our proposed Omni-CAD dataset is the only dataset available that simultaneously supports multi-view images, text, and point cloud conditioned data for CAD modeling. Notably, our dataset includes a large-scale collection of CAD models, second only to the ABC~\cite{ABC} and SketchGraphs~\cite{SketchGraphs} datasets. $\dagger$: SketchGraphs~\cite{SketchGraphs} focuses on the 2D CAD sketches instead of the 3D CAD models. $\ddagger$: Command Sequence Representation can convert to the B-rep representation.}
\label{tab:benchmark}
\end{table*}

On the other hand, multimodal large language models (MLLMs) have demonstrated their capability in content generation across different modalities ~\cite{guo2023pointbindpointllmaligning, ge2024worldgpt}. However, the use of MLLMs for CAD generation remains unexplored. While MLLMs support direct input from various modalities, meeting the requirements for conditional generation, there are two main challenges when applying MLLMs to conditional CAD generation: (1) the lack of an efficient representation that MLLMs can interpret and manipulate the CAD models, and (2) the unavailability of a large scale multimodal CAD dataset to align CAD models to the text, image, and point cloud modalities. To be specific, CAD models require a high level of specificity in terms of dimensions, connectivities, and functional requirements, while current LLMs are primarily trained in natural language. Thus, it is desirable to find a suitable CAD representation for LLM generation. Also, from the perspective of user-system interaction, how to bridge CAD models with text, image, and point cloud, these three modalities into a unified framework remains a significant challenge. Each of these modalities represents information in vastly different formats. The text describes concepts and attributes, images capture visual details, point clouds represent spatial data, and CAD models require precise geometric and structural definitions. From the dataset side, the current CAD datasets with command sequence, including the Fusion360~\cite{fusion360recon} and DeepCAD~\cite{DeepCAD}, are on a relatively small scale (8,625 and 179,133, respectively). A detailed comparison of datasets is provided in Tab.~\ref{tab:benchmark}. More importantly, current CAD datasets do not contain paired multimodal CAD data. To support the training of MLLMs, it is desirable to have an even larger scale dataset with CADs paired with different modalities to support the conditional CAD generation with MLLMs.

\begin{figure*}[ht]
\centering
\includegraphics[width=0.95\textwidth]{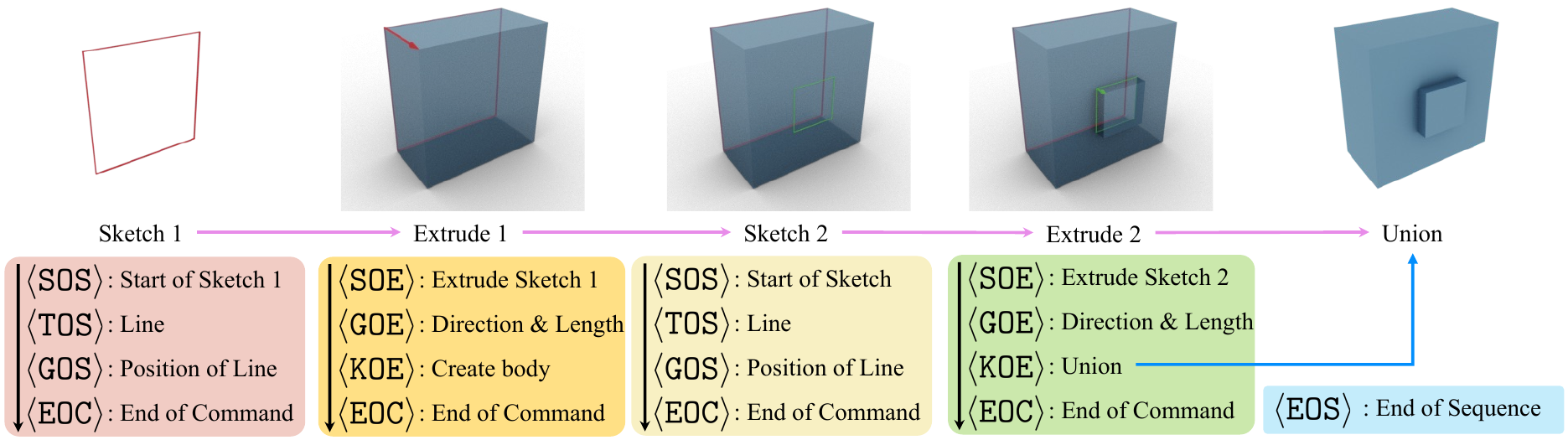}
\caption{A simple example about the construction process of a CAD model with command sequence representation. Starting with a \textit{sketch} operation on a chosen 2D plane, the \textit{extrusion} operation then ``drags" this 2D \textit{sketch} into a 3D solid volume. Further editing requires another \textit{extruded} 3D solid volume. Subsequently, the \textit{union} will ``merge" these two 3D solids into a single integrated solid. Other boolean operators from Constructive Solid Geometry (CSG) support the construction of more complex geometries. As a result, this CAD model can be represented with these command sequences.
}
\label{fig:command_sequence}
\end{figure*}

To address these challenges, we present CAD-MLLM to unleash the potential of MLLMs for CAD generation conditioned on multimodality inputs. Given that the primitive boundary representation of CAD models is non-sequential and unsuitable for an auto-regressive pipeline, motivated by the DeepCAD \cite{DeepCAD}, we instead utilize the command sequences, vectorizing them into a condensed sequential data flow that is more efficient for MLLMs to learn from. Combined with multimodality data, our model is capable of constructing complete CAD models by conditioning on text, images, point clouds, and any combination of them. When multiple modalities are input as a combination, our multimodal model demonstrates its strength by supplementing missing or suboptimal information from one data modality with input from another. To support our methodology, we propose a data annotation pipeline combined with a data augmentation method to generate a new multimodality-conditioned CAD dataset named Omni-CAD. Omni-CAD includes text descriptions, multi-view images, point clouds, and their corresponding constructive modeling command sequences. Omni-CAD reaches 453,220 models after data augmentation.

To evaluate the quality of the generated CAD models, although some previous CAD reconstruction works~\cite{parsenet, complexgen, point2cad, nvdnet} have proposed some well-established metrics for performance evaluation, such as utilizing sampled point clouds and patch topology to assess reconstruction fitting quality and patch structure fidelity, these metrics overlook an important nature of the CAD model: the overall topology quality of the CAD model in its final mesh representation. As a remedy, we propose three topology metrics, \textbf{Segment Error (SegE)}, \textbf{Dangling Edge Length (DangEL)}, and \textbf{Self-Intersection Ratio (SIR)}, to evaluate the topological quality of the final generated model. Additionally, since CAD models use boundary representation to form closed surfaces, we introduce \textbf{Flux Enclosure Error (FluxEE)} to quantify how well the generated model encloses space. These metrics are broadly applicable to general models in mesh representation as well. 

We conduct extensive evaluations on the proposed benchmark, and our experiments demonstrate that our method achieves state-of-the-art performance compared to other CAD generation methods and shows high robustness under various data flaws at the inference stage.

Our contributions can be summarized as follows: 
\noindent\begin{itemize}

\item We propose a unified multimodality-conditioned CAD generation method by leveraging the pretrained MLLM, and the condition can be text, images, point clouds, and any combination of these modalities. 

\item We present a data annotation pipeline and create a large-scale dataset, named Omni-CAD, the first multimodality CAD dataset includes constructive modeling command sequences and the corresponding textual descriptions, multi-view images, and point cloud data. 

\item We introduce four new evaluation metrics, namely, SegE, DangEL, SIR, and FluxEE, to evaluate the topological quality and enclosure of the generated CAD models, respectively. 

\item Extensive experiments show that our method demonstrates state-of-the-art performance over the baseline methods and high robustness under data flaws during inference.

\end{itemize}

\section{Related Work}
In this section, we will first review existing CAD generation methods based on different CAD presentations, namely, Boundary Representation (B-rep)~\cite{brep}, Constructive Solid Geometry (CSG)~\cite{CSG}, and Construction Command Sequence. Further, we will also review some MLLM related works.

\subsection{B-rep Based CAD Generation}

B-rep 3D models are depicted as graphs, incorporating both geometric primitives (e.g., parametric curves and surfaces) and topological primitives (e.g., vertices, edges, and faces) that trim and stitch surface patches to form solid models~\cite{brep}. Works about 
B-rep classification and segmentation have used various methods around the graph property, including graph neural networks~\cite{cadnet2020, solidgen, fusion360recon}, custom convolutions~\cite{lambourne2021brepnet}, and hierarchical graph structures~\cite{jones2023self, 10.1145/3478513.3480562, 10.1115/1.4063226}.

Some previous approaches have utilized predefined template curves and surfaces~\cite{smirnov2021patches, wang2022neural, wang2020pie, li2019supervised, parsenet} for B-rep generation. Specifically, PolyGen~\cite{nash2020polygen} uses pointer networks~\cite{NIPS2015_29921001} with Transformers~\cite{vaswani2017attention} to generate \textit{n-gon} meshes, a special case of B-rep models characterized by planar faces and linear edges. SolidGen~\cite{solidgen} and BrepGen~\cite{brepgen} can generate the entire B-rep models. SolidGen~\cite{solidgen} first synthesizes the vertices and then constructs them with the edge topology. BrepGen~\cite{brepgen} progressively denoises the faces, edges, and vertices utilizing Diffusion models~\cite{ho2020denoising}. Although B-rep is a direct representation of the boundary of the CAD model, it requires topological consistency, such as avoiding gaps and overlaps, inevitably introducing additional complexity to the CAD generation.

\subsection{CSG Based CAD Generation}
In CAD design, CSG is a widely-used technique for generating complex 3D shapes by combining solid primitives with boolean operators, like \textit{union}, \textit{intersection}, and \textit{subtraction}, to form a CSG tree finally. 
Recent CSG-based methods have concentrated on reconstructing 3D shapes as assemblies of primitives without relying on the ground truth CSG tree~\cite{kania2020ucsg, ren2021csg, yu2022capri, yu2023d2csgunsupervisedlearningcompact}. Meanwhile, CSG has been extensively leveraged in ``shape programs"~\cite{Neurosymbolic} with neural guidance~\cite{tian2018learning, NEURIPS2019_50d2d226, sharma2018csgnet} and without~\cite{10.1145/3272127.3275006, nandi2017programming, nandi2018functional, Gonzalez_2023}. Although the CSG tree can be converted into a B-rep model, the parametric CAD modeling~\cite{10.1016/j.cad.2016.01.003} with a sequence of 2D \textit{sketches} to be \textit{extruded} to 3D is still the primary paradigm for CAD designing and portable parametric editing.

\subsection{Command Sequence Based CAD Generation}

Recent available large-scale datasets~\cite{fusion360recon, DeepCAD} for parametric CAD modeling have facilitated the thriving of construction command sequence generation. Learning-based methods are investigated to utilize the history of the construction command sequence~\cite{fusion360recon, DeepCAD, skexgen, hnc, Khan_2024_CVPR} and the constraints of sketches~\cite{SketchGraphs} for generating engineering sketches and solid models. The generated sequences can be parsed using a solid modeling kernel to obtain an editable parametric CAD file. Furthermore, some works can generate the sequences or conduct reverse engineering conditioned on the sketching data~\cite{10.1145/3414685.3417807, seff2022vitruvion}, images~\cite{ganin2021computer, alam2024gencad}, 
voxel grids~\cite{10.1145/3550469.3555424}, point clouds~\cite{uy2022point2cyl} and target B-reps~\cite{xu2021inferring}
or without sequence guidance~\cite{ren2022extrudenet}. However, there is a notable absence of generation methods conditioned on text inputs, as well as those that handle more complex multimodal conditions. Additionally, a multimodal command sequence dataset for supporting advanced generation methods is lacking.

\subsection{Multimodal Alignment and Multimodal Large Language Models (MLLMs)}

Prior to MLLMs, many works, such as CLIP~\cite{radford2021learning}, have explored multimodal alignment. With the remarkable progress of LLMs~\cite{ChatGPT, achiam2023gpt,llama3modelcard, bi2024deepseek, bai2023qwen}, many efforts~\cite{zhu2023minigpt, wu2023next, liu2023visual, li2023blip} empowered the LLMs to the vision tasks by bridging with the pretrained visual encoders, and then downstream tasks have reached significant milestones. On this basis, some specialized models~\cite{wang2023drivemlm, shao2024lmdrive,wang2024mllm} in various vertical domains are being progressively explored. These tailored models aim to address specific challenges and enhance capabilities within each domain.

Meanwhile, some works explore the application of generating CAD models. The concurrent work Img2CAD~\cite{you2024img2cad} leverages VLMs to predict the global discrete structure and then conditioned on the structure, along with the semantics, to predict the continuous attributes. Another concurrent work Text2CAD~\cite{khan2024text2cadgeneratingsequentialcad} leverages both VLMs and LLMs for data annotation and uses Transformer~\cite{vaswani2017attention} structure to generate the full CAD sequence in an auto-regressive way. Some other recent works~\cite{alrashedy2024generatingcadcodevisionlanguage, wu2024cadvlmbridginglanguagevision, Yuan_2024} also investigate the utilization of VLM in CAD tasks. Compared to these works, our work supports not only image modality but also point and text modality simultaneously, and the MLLMs are directly empowered to predict the structure and attributes.

\newcommand{\cmdSOS}{\langle\mathtt{SOS}\rangle}
\newcommand{\cmdT}{\langle\mathtt{TOS}\rangle}
\newcommand{\cmdG}{\langle\mathtt{GOS}\rangle}
\newcommand{\cmdSOE}{\langle\mathtt{SOE}\rangle}
\newcommand{\cmdGOE}{\langle\mathtt{GOE}\rangle}
\newcommand{\cmdKOE}{\langle\mathtt{KOE}\rangle}
\newcommand{\cmdEOC}{\langle\mathtt{EOC}\rangle}
\newcommand{\cmdEOS}{\langle\mathtt{EOS}\rangle}

\section{Command Sequence Based CAD Representation}
\label{sec::command_sequence}
 
From a user-interaction perspective, the popular industrial standard for the creation of CAD models can be described as the sequence of operations performed by CAD software (e.g. OpenCascade~\cite{opencascade}, Fusion360~\cite{fusion360}, and Solidworks~\cite{systemes2011solidworks}). To create a solid shape, a user first needs to create a closed curve profile as a 2D \textit{sketch}, and then \textit{extrude} it into a 3D solid shape. To further create complex surfaces or objects, CSG~\cite{CSG} enables the user to combine simpler objects by applying boolean operators, such as \textit{union}, \textit{intersection}, and \textit{subtraction}, which allows for the generation of visually intricate objects through the combination of a few primitive shapes. 

Given a CAD command sequence, it can be automatically transformed into a B-rep representation of a CAD model through a CAD modeling library, like PythonOCC~\cite{Paviot2022}. Following DeepCAD~\cite{DeepCAD}, we represent a sketch-and-extrude CAD model using a sequence with five types of tokens, \textbf{Start and End-command token}, \textbf{Topology token}, \textbf{Geometry token}, \textbf{Kind-of-extrusion token}, \textbf{End-of-sequence token}, which are aligned with notation in Tab.~\ref{tab:notation}:

\begin{figure*}[h]
\centering
\includegraphics[width=0.95\textwidth]{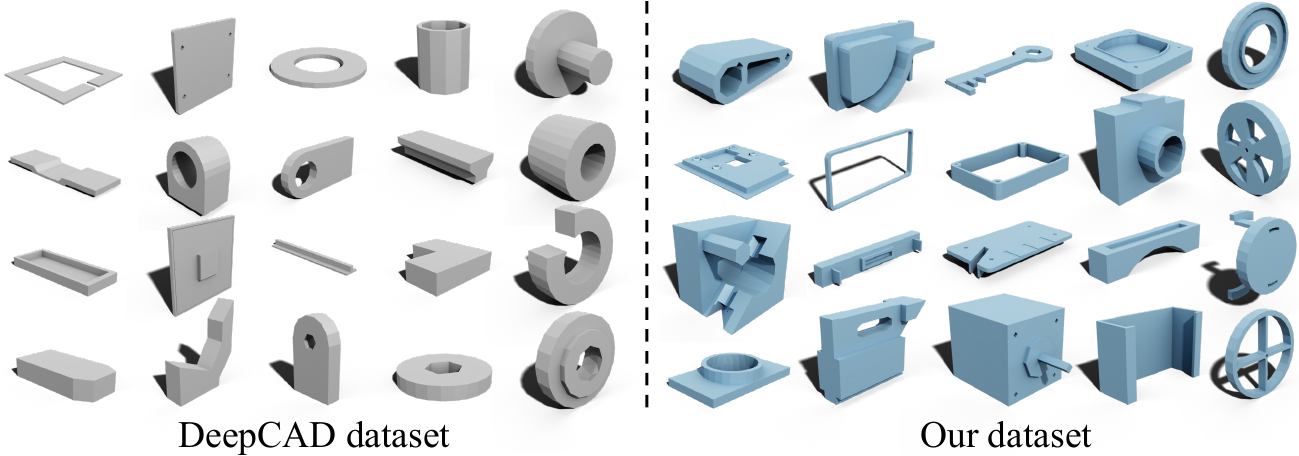}
\caption{Qualitative comparison between our CAD command sequence dataset and DeepCAD~\cite{DeepCAD} dataset. According to Sec.~\ref{sec:create_dataset}, DeepCAD dataset is part of our created dataset. In the visualization of our dataset, we \textbf{exclude} the CAD models' IDs that have been included in the DeepCAD dataset. The extension part of our dataset contains more complex and realistic models with more details. \textbf{Best viewed zoomed in.}
}
\label{fig:dataset_gallery}
\end{figure*}

As an example, a CAD model can be constructed with the command sequence through the \textit{union} of two \textit{extruded} solid shapes, which is illustrated in Fig.~\ref{fig:command_sequence}. 

Below, we define the \textit{sketch} and \textit{extrusion} operations in the format of commands consisting of various continuous attributes in a practical way.

\textbf{Sketch.} According to the CAD terminology, a \textit{profile} is a closed region consisting of one or several loops, where the curve commands in each loop are concatenated. Thus, except the start point of the first curve is the origin of the plane, the remaining curves' start points are the endpoints of the predecessor curves typically. In practice, for the \textit{$\langle\mathtt{TOS}\rangle$}, we consider the most three types of curve commands, \textit{lines} ($L$), \textit{arcs} ($A$), and \textit{circles} ($R$). 

The corresponding \textit{$\langle\mathtt{GOS}\rangle$} geometry of each type of \textit{$\langle\mathtt{TOS}\rangle$} curve is defined as follows:

\begin{itemize}
    \item $L : (x, y)$, where $(x, y)$ defines the endpoint of a line.
    \item $A: (x, y, \alpha, f)$ which defines an arc with the endpoint $(x, y)$ and sweep angle $\alpha$. $f$ refers to the counter-clockwise flag.
    \item $R: (x, y, r)$, where $(x, y)$ is the center of an circle with a radius $r$. 
\end{itemize}

\textbf{Extrusion.} As mentioned above, the extrusion command serves a dual purpose; it needs to provide the information not only on how to transform a sketch profile into a 3D shape by extending it along a specified path but also on the spatial relationship and merge operation of the newly formed 3D shape with other existing 3D shapes to form the final 3D shape. Therefore, the extrusion command can be defined as $E:(\theta, \phi, \gamma, x, y, z, s, e_{p}, e_{n}, b, u)$, where $(\theta, \phi, \gamma)$ are the three Euler angles determining the extrusion orientation, $(x, y, z)$ refers to the origin of the sketch plane, $s$ represents the scale factor. Besides that, $e_{p}, e_{n}$ denotes the extrusion distance towards the positive direction and negative direction respectively. The parameters related to the geometry of \textit{extrusion} operation form \textit{$\langle\mathtt{GOE}\rangle$}. Additionally, $b$ and $u$ are two type arguments specifying the volume boolean type (e.g. \textit{joining}, \textit{intersecting}, \textit{cutting}) and extrusion type (e.g. \textit{one-sided}, \textit{symmetric}, \textit{two-sided}), which correspond to \textit{$\langle\mathtt{KOE}\rangle$}. 

By integrating these two operations, each sequence can be vectorized as 16 distinct variables. To save the length of the command sequence without affecting the information of the command, instead of setting unused parameters in the command sequences to be -1 as DeepCAD~\cite{DeepCAD}, we use a particular Place Holder Token, combining with other tokens acting as the End-of-command token, \textit{$\langle\mathtt{EOC}\rangle$}. Specifically, when the last few variables of a sequence are all the placeholder tokens, these placeholder tokens will act as an \textit{$\langle\mathtt{EOS}\rangle$} to indicate the end of the current command.

\begin{table}[htb]
\begin{tabular}{|p{0.25in}|p{2.75in}|}
\hline
\multicolumn{2}{|l|}{\textbf{Notation}}\\ \hline

\multicolumn{2}{|l|}{\textbf{Sketch Related Token:}}\\

$\cmdSOS$ & 
\textbf{Start-of-sketch token}: Denotes the start of a \textit{sketch} operation.\\ 

$\cmdT$ & 
\textbf{Topology-of-sketch token}: Specifies the type of curve used in the \textit{sketch} operation. This token indicates whether the curve is a line, arc, or circle.\\ 

$\cmdG$ & 
\textbf{Geometry-of-sketch token}: Contains the coordinates of points and geometric parameters that define the shape of the sketch. These tokens provide the necessary geometric information for constructing the curves. Note that every model is normalized within a cube range from $[-1, 1]^3$ before being quantized to $256$ levels.\\

\hline
\multicolumn{2}{|l|}{\textbf{Extrude Related Token:}}\\

$\cmdSOE$ & 
\textbf{Start-of-extrusion token}: Denotes the start of a \textit{extrusion} operation.\\ 

$\cmdGOE$ & 
\textbf{Geometry-of-extrusion token}:  Contains parameters related to the \textit{extrusion} process. These parameters include the direction, type, and length of the extrusion.\\

$\cmdKOE$ &
\textbf{Kind-of-extrusion token}: Identify the associated volume boolean operations after creating the solid of this \textit{extrusion} operation. The volume boolean operations include \textit{union}, \textit{intersection}, or \textit{subtraction} with the current CAD design, which will generate a more complex solid.\\ 

\hline
\multicolumn{2}{|l|}{\textbf{Ending Related Tokens:}}\\

$\cmdEOC$ & 
\textbf{End-of-command token}: Denotes the end of a operation command.\\

$\cmdEOS$ & 
\textbf{End-of-sequence token}: Denotes the end of the entire command sequence.\\

\hline
\end{tabular}
\caption{Notation for sequential tokens. This table details the essential elements for constructing a sketch command, an extrusion command, and an entire command sequence.}
\label{tab:notation}
\end{table}

\begin{figure}[h]
\centering
\begin{subfigure}[b]{0.45\textwidth}
    \centering
    \includegraphics[width=\textwidth]{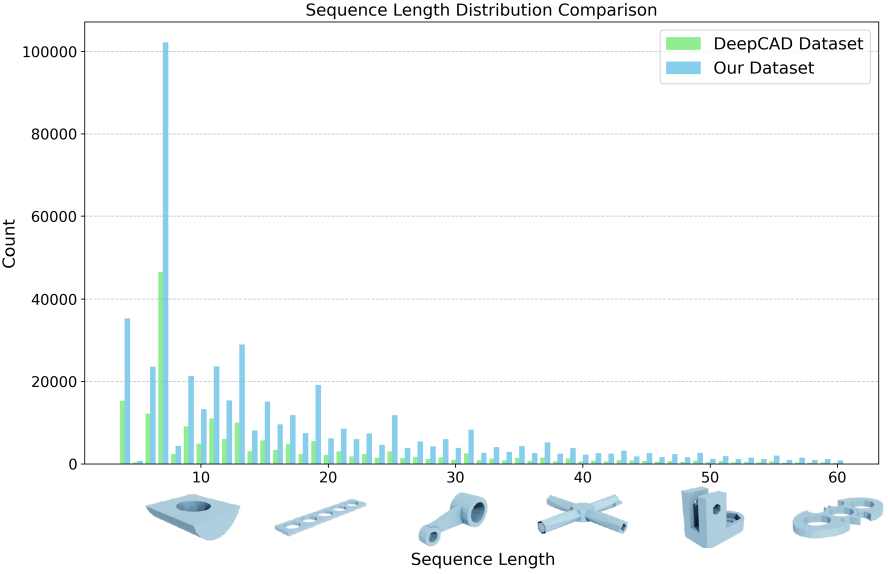}
    \caption{The statistical comparison between our dataset and DeepCAD~\cite{DeepCAD} dataset over the command sequence length per CAD model. The longer sequence length indicates the more complicated case.}
    \label{fig:sequence_length_statistic}
\end{subfigure}
\vfill
\begin{subfigure}[b]{0.45\textwidth}
    \centering
    \includegraphics[width=\textwidth]{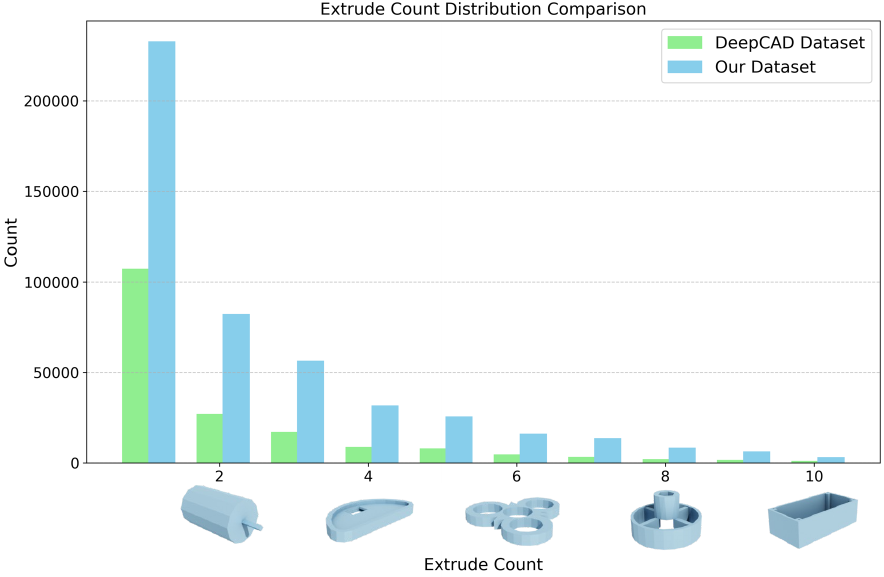}
    \caption{The statistical comparison between our dataset and DeepCAD~\cite{DeepCAD} dataset over the \textit{extrusion} operation counts per CAD model.
    The more \textit{extrusion} operation count indicates the more challenging case. \textbf{Best viewed zoomed in.}
    }
    \label{fig:extrude_count_statistic}
\end{subfigure}
\caption{The statistical comparison between our dataset and DeepCAD~\cite{DeepCAD} dataset. The statistics are conducted \textbf{before data augmentation}. 
The charts indicate that our dataset extends the data over a wide range of sequence counts and \textit{extrusion} operation counts with more challenging cases. 
}
\label{fig:combined_statistic}
\end{figure}

\section{Creation of Our Large-Scale Multimodal CAD Dataset}
\label{sec:create_dataset}

Several datasets for CAD modeling are publicly accessible. The ABC dataset~\cite{ABC} comprises 1 million CAD designs sourced from Onshape~\cite{onshape}, a cloud-based platform for product design. However, these CAD designs are initially provided in B-rep form, which lacks the detailed information needed to recover the construction operations. In contrast, the Fusion360 Reconstruction Dataset~\cite{fusion360recon} offers CAD modeling sequences created by human designers. Despite this, the dataset contains only $8,625$ CAD designs, which is insufficient for training a generalized generative model. Apart from the insufficient scale of the datasets, current datasets only provide information related to CAD models. To lower users' barriers in creating CAD models and enable non-experts to bring their ideas to life through arbitrary multimodal conditions, a dataset with corresponding textual descriptions, multi-view images, and point cloud data alongside CAD models is essential. However, such a dataset does not currently exist. 

Therefore, we create a new large-scale multimodal CAD dataset that simultaneously provides CAD command sequences and corresponding data in three modalities, which we hope will inspire and accelerate advancements in future research. 

\subsection{CAD Command Sequences Generation and Augmentation}

Originating from the ABC dataset~\cite{ABC}, we adopt DeepCAD's~\cite{DeepCAD} approach to process CAD designs, utilizing Onshape’s developer API~\cite{onshapeDev} and parsing with Onshape's FeatureScript~\cite{onshapeFeatureScript}. 

 As shown in Sec.~\ref{sec::command_sequence}, the command sequence representation method focuses on the 2D plane and the process to transform into a 3D shape body, not including edge or face primitives that are required by some specific commands, such as \textit{chamfer} and \textit{fillet}. In CAD modeling, the \textit{chamfer} and \textit{fillet} operations are commonly employed to mitigate sharp edges and corners in engineering and design contexts. In detail, a \textit{chamfer} replaces a sharp directional change with an angled slope, whereas a \textit{fillet} introduces a smooth, curved transition between two surfaces. Due to the vectorized sequence representation limitation, this parsing method would remove all CAD designs containing \textit{chamfer} or \textit{fillet} operations.

However, unlike DeepCAD, which directly removes all the designs with any of these two operations, we individually remove each \textit{chamfer} and \textit{fillet} operation and retain the CAD design if it maintains a complete topology. As a result, we initially collect a dataset of $275,717$ models, nearly 1.54$\times$ the $179,133$ designs reserved by DeepCAD.

We additionally augment our data by extracting intermediate CAD designs after each \textit{extrusion} operation. For example, a CAD design with 7 \textit{extrusion} operations can be augmented into 7 CAD designs. In the end, we collect a total of $453,220$ augmented CAD command sequence data. Note that to ensure fair testing and prevent the augmented data's interrelations from providing undue advantages, we divide the dataset into training and testing sets before we apply the data augmentation strategy exclusively to the training set.

Fig.~\ref{fig:dataset_gallery} visualizes the qualitative comparison of our dataset and DeepCAD~\cite{DeepCAD} dataset. The statistics of our extension in both challenging sequence length and challenging \textit{extrusion} operation count can be observed from Fig.~\ref{fig:combined_statistic}.

\begin{figure*}[h]
\centering
\includegraphics[width=0.95\textwidth]{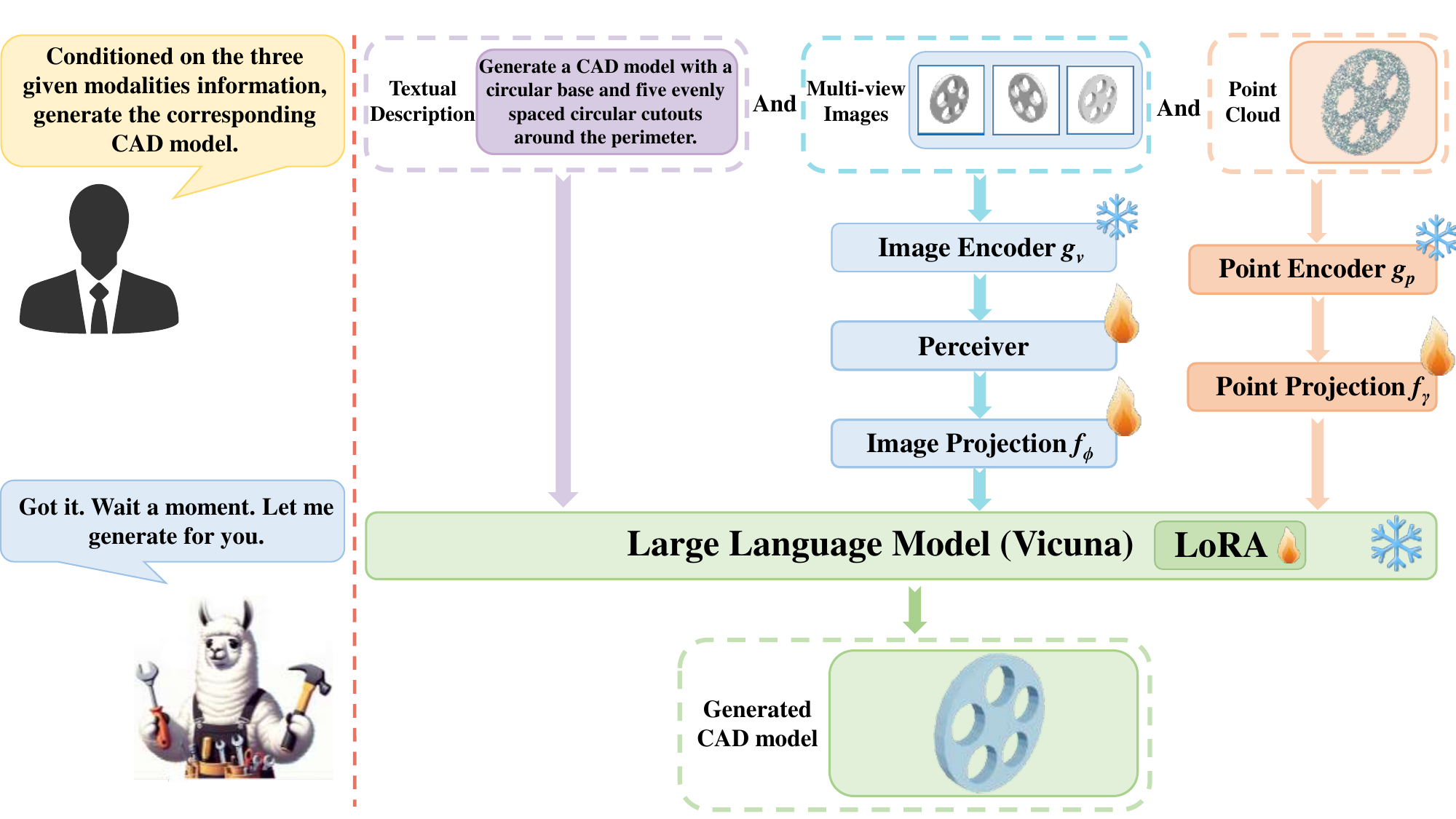}
\caption{\textbf{Our network architecture}. {The network could process three single modalities of information of input or any combinations of them, each uniquely color-coded. We consider the most complex combination of modalities, where three different inputs are provided simultaneously. Except for the textual descriptions, each modality is first processed through its corresponding frozen encoder before being further integrated. Subsequently, they are passed through a trainable projection layer, aligning them within a unified language feature space. The fine-tuned Large Language Models (LLMs), augmented with Low-Rank Adaptation (LoRA), then process a combination of the prompt and the projected embeddings, enabling the accurate generation of CAD models.}}
\label{fig:pipeline}
\end{figure*}

\subsection{Conditional Data Generation}
In addition to generating the vectorized command sequence representations for CAD models, our more important objective is to address the current gap in datasets that lack multimodal information corresponding to CAD models, such as images, point clouds, and textual descriptions. 

For each CAD model, we render multi-view images from eight fixed perspectives. For the point cloud data, we randomly sample points and record their corresponding normal information.

For the textual description of each CAD design, since there is currently no effective method to directly input a CAD representation into an MLLM and achieve good caption results, we use previously rendered multi-view images as input for the MLLM. Due to budget constraints and considering the quality of generated textual descriptions, the inference speed of the MLLM, and whether the model supports multiple images as input, we leverage the open-sourced MLLM InternVL2-26B~\cite{chen2024far, chen2024internvl} and randomly select four view images as input for each CAD design to generate high-quality text captions. The prompt is as follows: “\textit{These are the rendering images from 4 views of a CAD model. Please describe these images with one caption, and mainly focus on the shape and appearance of the foreground while ignoring the details of the background.}”. To standardize the format of the textual descriptions, we include format constraints in the prompt, requiring all outputs to begin with “\textit{Generate a CAD design with }”. Some examples of conditioned data are provided in the supplementary.

\section{Method}

Besides the general CAD model's representation B-rep, recent works~\cite{DeepCAD, skexgen, hnc, seff2022vitruvion} show CAD command sequences are able to utilize the history of CAD modeling sequences and constraints on the sketches. We present our CAD-MLLM, an MLLM model tailored for CAD generation based on modeling sequences. CAD-MLLM supports cross-modal inputs, including text, image, and point cloud, as conditions for generating novel CAD models.

\subsection{CAD-MLLM Architecture}
\label{sec:architecture}
As shown in Fig.~\ref{fig:pipeline}, the proposed CAD-MLLM consists of three modules: visual data alignment, point data alignment, and the large language model. Notably, as text input data is directly fed into the LLM for embedding extraction, there is no need for an additional text alignment module. 

\begin{itemize}
    \item \textbf{Visual Data Alignment.}\\
     Given the input multi-view images $X_{v} = \{X_{v}^{1}, X_{v}^{2}, \dots, X_{v}^{k}\}$ where $k$ specifies the number of views, the vision encoder $g_{v}$ extracts independent visual features from each image. These features are then concatenated into a unified representation $H_{v} \in \mathbb{R}^{k \times (1 + L_s) \times d_s}$, where $1 + L_s$ denotes the length of tokens, which includes the class head token as well as the patch tokens, and $d_s$ is the dimension. Drawing inspiration from previous perceiver-based transformer architectures~\cite{jaegle2021perceiver}, we implement a cross-attention layer to integrate the information from the $k$ multi-view images information contained in the input $H_{v}$ into a learnable query token $Q \in \mathbb{R}^{1 \times L_q \times d_q}$, where $L_q$ and $d_q$ is the length and dimension of token $Q$. Additionally, an image projection layer $f_{\phi}$ is employed to project the visual signals into the feature space of the pretrained LLM where $\phi$ denotes the parameters to be learned in the projection layer.
    \begin{equation}
    \begin{aligned}
    H_{v} &= \text{Con}(g_{v}(X_{v})) \\
    E_{v} &= f_{\phi}(\text{CA}(Q, H_{v}))
    \end{aligned}
    \end{equation}
    
    \item \textbf{Point Data Alignment.}\\
    Similar to visual inputs, when provided with point cloud data $X_{p}$, a point encoder $g_{p}$ is used to extract features. These features are then projected into a feature space comprehensible by the LLM through a linear layer $f_{\gamma}$ where $\gamma$ denotes the parameters to be learned in the point projection layer.
    \begin{equation}
    E_{p} = f_{\gamma}(g_{p}(X_{p}))
\end{equation}

    \item \textbf{LoRA based Large Language Model Finetuning.}\\
    Large language models serve a dual purpose in our approach. On one hand, for the textual description $X_{l}$ of the input CAD model, we follow Vicuna's method~\cite{llama3modelcard, vicuna, touvron2023llama, touvron2023llama2} by utilizing a BPE tokenizer~\cite{sennrich2015neural} to obtain text embeddings $E_{l}$. On the other hand, we input the concatenated features of the conditioned modalities into the large language model, which is tasked with predicting the sequence of commands for the CAD model as the output. To optimize our model while minimizing the number of learnable parameters, we implement Low-Rank Adaptation (LoRA) ~\cite{hu2021lora} to fine-tune an open-sourced LLM (Vicuna-7B \cite{vicuna}), parameterized by $\delta$.

\end{itemize}

\subsection{Training Objective}
\label{sec:objective}
We leverage the pretrained visual encoder $g_{v}$ and point encoder $g_{p}$ and keep them frozen. The overall objective, as shown in Fig.~\ref{fig:pipeline}, is to train the visual perceiver, image projection layer $f_{\phi}$, point projection layer $f_{\gamma}$ and the LoRA $\delta$. We denote the trainable parameters as $\theta = \{\eta, \phi, \gamma, \delta \}$ where $\eta$ is the parameter for the visual perceiver.

Inspired by~\cite{ge2024worldgpt, bengio2009curriculum}, we adopt a curriculum-based progressive training strategy, gradually introducing modalities in the following order: textual descriptions, point clouds, and multi-view images. The newly introduced modalities are randomly combined with existing ones to form various multimodal input configurations, allowing for comprehensive training across diverse input scenarios. 

Considering the most complex case, when the user inputs the text description $X_{l}$, multi-view images $X_{v}$  and point cloud data $X_{p}$ as condition at the same time, as mentioned before, the visual data alignment module extracts the image feature $E_{v}$, point data alignment module extracts the point embedding $E_{p}$, and the textual embedding $E_{l}$ is obtained by the BPE tokenizer. The language modeling (LM) loss is adopted to supervise the training of CAD-MLLM:
\begin{gather}
    \mathcal{L}_{\text{LM}}=-\sum_{t=1}^L \log P_{\theta}(y_{i,t}| y_{i,<t}, E_{v}, E_{p}, E_{l}).
    \label{eq.ntp_loss}
\end{gather}
where $y_{i}= \{y_{i,1}, y_{i,2}, \dots, y_{i,L}\}$ is the predicted response sequence with length $L$ for the $i^{th}$ input.

\section{Experiments}

\subsection{Experimental Setup}

\subsubsection{Datasets}

We use our multimodal CAD dataset for training and evaluation, which involves $453,220$ command sequences that are vectorized into the specific data flow we use. It also contains multimodal data (text/multi-view images/point clouds) for each vectorized augmented data for our multimodal training. We divide our Omni-CAD dataset into training and testing sets in a 9:1 ratio, with $425,726$ pairs of data used for training and $27,494$ for evaluation.

\subsubsection{Training Details}

We implement CAD-MLLM with PyTorch~\cite{pytorch} and train it across 16 NVIDIA H800 80G GPUs for 20 epochs, taking approximately 47 hours. We employ an AdamW~\cite{adamw} optimizer with a learning rate 2e-5 and a linear decay. The dropout rate is set to 0.1, and the batch size is 8192, using a micro-batch size of 1 and 512 gradient accumulation steps. The maximum sequence length is 1024. For the large language model component, we utilize Vicuna-7B \cite{vicuna}. The DINO v2\cite{oquab2023dinov2, darcet2023vitneedreg} is used as the visual encoder, and Michelangelo~\cite{zhao2023michelangelo} is used as the point cloud encoder. Due to computational resource limitations, particularly when handling the most complex multimodal inputs, we limit the number of multi-view images to 2 in this work. As mentioned in Section~\ref{sec:objective}, a curriculum-based progressive training strategy is introduced during the training process. When given a batch of data, it is essential to pre-determine the modality information carried by each data sample. This modality selection is made randomly and with equal probability from the available combinations for the current phase. As a result, the chosen modalities for each data sample can vary, potentially consisting of either a single modality or a combination of multiple modalities. Notably, for the image inputs, since each CAD model is rendered from eight different viewpoints, consequently, two images are randomly sampled from these eight rendered views to serve as input. 

\subsubsection{Baselines}
\label{sec:baseline}

Our method can generate CAD command sequences with multimodal conditions. 
Therefore, we conduct our experiments on different tasks.

\noindent \textbf{Point Clouds Conditioned CAD Generation}: Since the point clouds condition offers a precise 3D reference for the target CAD model, we assess the reconstruction capability of our generation model against several baseline methods, including two different kinds of techniques: ``B-rep"-based reconstruction and ``command sequence"-based generation:

\begin{enumerate}[leftmargin=*]
    \item \textbf{``B-rep"-based reconstruction baselines}: We compare our method with ``B-rep"-based point clouds CAD reconstruction baselines ParSeNet~\cite{parsenet}, ComplexGen~\cite{complexgen}, Point2CAD~\cite{point2cad} and NVDNet~\cite{nvdnet}. Notice that these reconstruction methods target the B-rep reconstruction of the CAD models, which is different from the CAD command sequence. In the following comparisons, we can roughly consider the conditional generation task of our method based on point clouds as the point cloud reconstruction task.
   
    \item \textbf{``Command Sequence"-based generation baseline}: We additionally compare our method with ``Command Sequence"-based CAD generation baseline DeepCAD~\cite{DeepCAD} on point clouds reconstruction. We conduct the point clouds conditioned DeepCAD with official implementation and with our dataset, using PointNet++~\cite{pointnet++} to encode and embed the point cloud to the latent vector. 

\end{enumerate}

\noindent \textbf{Image Conditioned CAD Generation}: To the best of our knowledge, currently, there is no open-sourced ``image-to-CAD" baseline to compare. Instead, we compare with the ``image-to-mesh" baselines. As mentioned in Sec.~\ref{sec:architecture}, our multimodal model is able to generate a CAD model with multi-view images as a condition. We select the methods that support the multi-view images as a condition, including the InstantMesh~\cite{xu2024instantmesh} and SpaRP~\cite{xu2024sparp}, for performance comparison.

\noindent \textbf{Text Conditioned CAD Generation}: To the best of our knowledge, currently, there is no open-sourced ``text-to-CAD" baseline to compare. Instead, we compare with the ``text-to-mesh" baselines, Michelangelo~\cite{zhao2023michelangelo} and Tripo~\cite{tripo3d_ai}.

\begin{figure*}[h]
\centering
\includegraphics[width=0.95\textwidth]{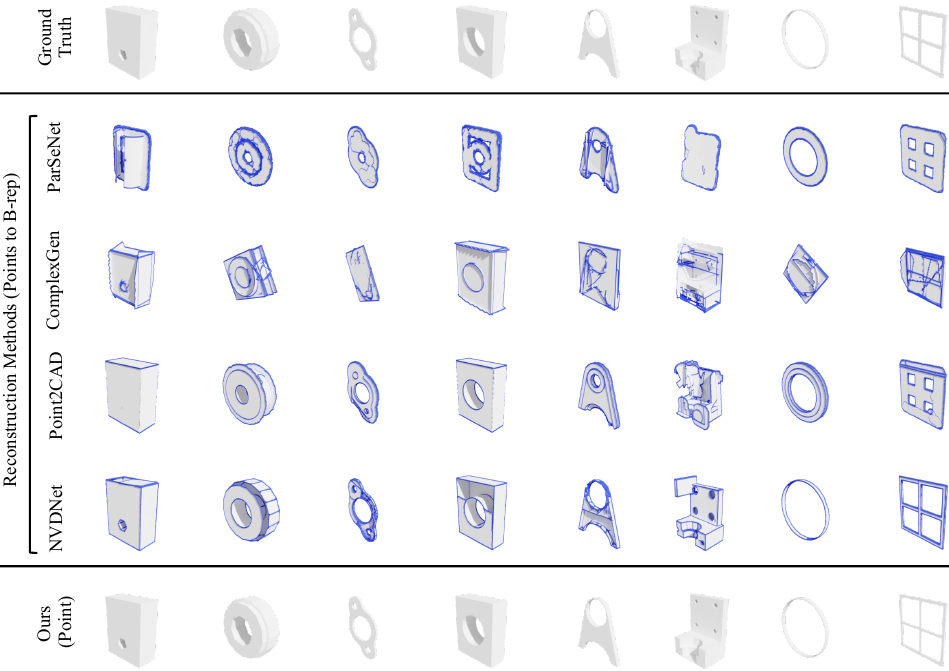}
\caption{We present qualitative point-based reconstruction results on our dataset and compare our generative method with the point-based B-rep reconstruction baseline. Blue lines highlight the dangling edges in the reconstructed model. Our method produces high-fidelity reconstructed results. Most of our reconstructed results are strict manifolds and do not have dangling edges (do not have blue lines). The results of the comparison of reconstruction baselines show that they have lots of dangling edges. This figure illustrates that our method outperforms from the topological aspect.}
\label{fig:recon_exp2}
\end{figure*}

\begin{table*}[h]
\setlength{\tabcolsep}{4pt} %

\centering
\resizebox{\linewidth}{!}{
\begin{tabular}{cc|ccccccc}
\toprule
\multicolumn{2}{c|}{Point-based CAD Reconstruction}      & Chamfer($\times 100$)$\downarrow$        & F-score($\times 100$)$\uparrow$  & Normal C($\times 100$)$\uparrow$ & SegE$\downarrow$ & DangEL$\downarrow$ & SIR(\%)$\downarrow$   &   FluxEE($\times 100$)$\downarrow$        \\ \hline
\multicolumn{1}{c|}{\multirow{4}{*}{Reconstruction Methods}} 
& ParSeNet~\cite{parsenet} & 4.59          & 42.56          & 46.83   & 10.92 & 78.82 & 13.87  &  398.524  \\
\multicolumn{1}{c|}{}   & ComplexGen~\cite{complexgen} & 1.65        & 86.12          & 64.82   & 17.72  &  55.32 &  22.57  & 115.918  \\
\multicolumn{1}{c|}{}   & Point2CAD~\cite{point2cad}    & 1.25 & 89.85 & 65.90  & 15.82 & 9.23 & 11.38 & 97.453 \\ 
\multicolumn{1}{c|}{}   & NVDNet~\cite{nvdnet}    & \textbf{0.82} & \textbf{98.94} & \textbf{93.94} & 37.22 & 47.97 & 2.60 & 14.550\\
\hline
\multicolumn{1}{c|}{\multirow{2}{*}{Generation Methods}} & DeepCAD(Point)~\cite{DeepCAD}  & 4.51    & 71.83   & 63.68   & 8.98  & 1.26 & 5.73  &  0.347  \\
\multicolumn{1}{c|}{} & Ours(Point)    & \textbf{1.85} & \textbf{90.88} & \textbf{79.71} & \textbf{1.66} &  \textbf{0.46} &  \textbf{1.31}  & \textbf{0.044}  \\

\bottomrule
\end{tabular}
}
\caption{The quantitative results on point-based reconstruction tasks. Our method's performance is comparable to some ``B-rep"-based reconstruction methods in ``point cloud"-based reconstruction metrics (Chamfer, F-score, Normal C). However, it significantly outperforms these methods in our proposed topological metrics (SegE, DangEL, and SIR) and the enclosure metric (FluxEE). Moreover, our method consistently surpasses the command sequence-based generation baseline across all evaluated metrics.}
\label{table:point_recon}
\end{table*}

\subsubsection{Evaluation Metrics}

We follow the metrics in existing work for \textbf{CAD reconstruction}, and we additionally propose four new metrics covering the aspect of \textbf{CAD topology} and \textbf{model enclosure} to quantify the quality of the generated models better.

\noindent \textbf{CAD reconstruction metrics}:

Following previous reconstruction methods' evaluation~\cite{nvdnet, point2cad}, we compare the Chamfer Distance (Chamfer) and F-score with a $0.05$ threshold. Additionally, we compare the Normal Consistency (Normal C) between the ground truth model and the reconstructed/generated model following the evaluation of ~\cite{Yu2022MonoSDF, debsdf2024}. Note that the objects are normalized to $[-0.5, 0.5]^3$ for reconstruction evaluation.

\noindent \textbf{CAD topology metrics}:

The reconstruction metrics mentioned above ignore the reconstructed model's topology quality and only focus on the point cloud. This oversight neglects critical topological information and fine-grained details inherent in CAD representations. 
Though Complexgen~\cite{complexgen} proposed patch-to-patch topology accuracy metrics for measuring structural fidelity, it just evaluates patch-to-patch topology and ignores the overall topology quality as an assembled CAD model in mesh representation. To address this, we propose three additional metrics to better evaluate the generated CAD model's topology quality. We clarify that we treat edges in the mesh representation as connectivity descriptions. We define \textit{"two nodes in the same segment"} as an edge connecting them. Additionally, we wish to prevent non-manifolds from being used for the reconstructed models. We call the edges that are only bounded by one face, as \textit{"dangling edges"}, which will lead to a non-manifold structure. GeometryCentral~\cite{geometrycentral} and CGAL~\cite{cgal:eb-24a} are used in our implementation.

\begin{enumerate}[leftmargin=*]
    \item \textbf{Segment Error (SegE)} measures the fidelity of the topology from the segment aspect. We denote $S(\cdot)$ as the segment number among all nodes in a mesh. 
    
    The SegE of the CAD model is defined as follows: 

    \begin{equation}
        \text{SegE}(\hat{G})= \cfrac{|S(\hat{G}) - S(G)|}{S(G)}
    \end{equation}

    where $G$ is the ground truth model and $\hat{G}$ is the generated model.
   
    \item \textbf{Dangling Edge Length (DangEL)} measures the quantity of the non-manifold structure. For arbitrary mesh, the dangling edges are the edges that are only bounded by one face, which harms the manifold structure. We locate the dangling edges by executing a half-edge~\cite{MULLER1978217, DDG} traversal over the whole mesh and then detecting the edges only accessed once. Finally, we sum up the length of all dangling edges in a mesh as the evaluation metric.
    
    \item \textbf{Self-Intersection Ratio (SIR)} measures the ratio of the self-intersected faces among all faces. A mesh with self-intersections also does not meet the requirements of a manifold. We compute the number of self-intersected faces and then divide the total number of faces.
    
\end{enumerate}

\noindent \textbf{Model enclosure metric}:

Additionally, the CAD models utilize boundary representation to form closed surfaces. Besides evaluating topology quality, the enclosure of the generated models is also an important aspect to consider. For a general continuous closed surface, according to the Gauss's Divergence Theorem~\cite{Gauss1877}, for a vector field $\mathbf{F}$, the flux through a closed surface is equal to the volume integral of the divergence of $\mathbf{F}$ over the region enclosed by the surface.

\begin{equation}
\oint_S \mathbf{F} \cdot \mathbf{n} \, dS = \int_V \nabla \cdot \mathbf{F} \, dV
\label{eq:gaussian_divergence}
\end{equation}
where $S$ is the closed surface, $V$ represents the volume enclosed by $S$, $dS$ represents the surface differential, $\mathbf{n}$ represents the outward-pointing unit normal vector of the surface differential, $dV$ represents the volume differential.

For simplicity, we define $\mathbf{F}$ as a constant vector field $(1,1,1)$, since the divergence of a constant vector field is always zero.
\begin{equation}
\oint_S \mathbf{F} \cdot \mathbf{n} \, dS = \oint_S (n_x + n_y + n_z) dS = \int_V \nabla \cdot \mathbf{F} \, dV = 0
\label{eq:flux_divergence}
\end{equation}
where $n_x$, $n_y$, and $n_z$ represent the components along the $x$, $y$, $z$ axes of the unit normal vector $\mathbf{n}$.

For the discrete computation, the discretization of an ideal closed surface suffices the following:
\begin{equation}
\oint_S (n_x + n_y + n_z) \, dS \approx \sum_{i=1}^N (n_{i,x} + n_{i,y} + n_{i,z}) \, dS_i
\label{eq:flux_discrete}
\end{equation}
where $N$ is the total number of discrete meshes, and $dS_i$ is the area of the $i$-th discrete mesh element. $n_{i,x}$ represents the $x$-component of $\mathbf{n}$ at the $i$-th discrete mesh element, with similar definitions for other axes.

For an ideal closed surface in discrete form, the discrete integral approximation becomes
\begin{equation}
\sum_{i=1}^N (n_{i,x} + n_{i,y} + n_{i,z}) \, dS_i = 0
\label{eq:flux_discrete_approx}
\end{equation}

The enormous flux of this constant vector field through the discrete mesh indicates that this mesh is far from the ideal closed surface, which is not expected for our generated model. We define additional metrics as:

\begin{enumerate}[leftmargin=*]
    \setcounter{enumi}{3}
    \item \textbf{Flux Enclosure Error (FluxEE)} measures the degree of enclosure for the mesh. The FluxEE of the CAD model is defined as follows:

    \begin{equation}
        \text{FluxEE}(\hat{G})= \left| \sum_{i=1}^N (n_{i,x} + n_{i,y} + n_{i,z}) \, dS_i \right|
        \label{eq:FE}
    \end{equation}

\end{enumerate}

\subsection{Results}

\subsubsection{Point Conditioned CAD Generation}
\label{sec:recon_exp}

We conduct our training for this point-based CAD reconstruction experiment using only point clouds as input, aligning our methodology with that of the selected baselines, to ensure a fair comparison.

Tab.~\ref{table:point_recon} presents a comparison of our method against the aforementioned baselines, focusing on reconstruction metrics as well as the new topology and enclosure metrics we proposed. Notably, our method demonstrates superior performance on reconstruction metrics, even outperforming some reconstruction methods and trailing only behind NVDNet~\cite{nvdnet}. This gap can be attributed to its Voronoi cells splitting and primitive fitting design. However, in terms of topology and enclosure metrics, our method significantly outperforms the baselines. From the visual comparison in Fig.\ref{fig:recon_exp2}, we can also see that our method generates high-fidelity CAD models. While the baselines exhibit numerous dangling edges in their reconstructions, as indicated by the blue lines in the figure, most results of our approach exhibit strict manifold structures without any dangling edges, showcasing superior topological quality. The good topology of our results benefits from the accuracy of our generated command sequences.

\begin{figure}[h]
\centering
\includegraphics[width=0.45\textwidth]{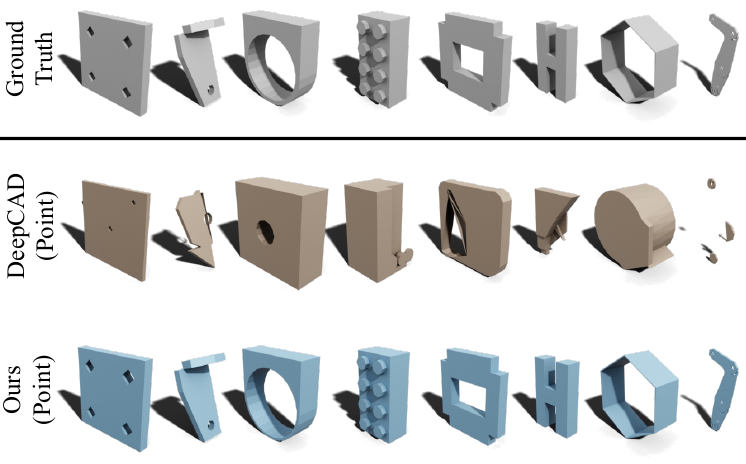}
\caption{
We qualitatively compare point-based reconstruction results on our dataset with the baseline generative method. Our method successfully generates the correct command sequence with the corresponding point cloud condition. 
}
\label{fig:recon_exp2_gen}
\end{figure}

When compared to the generative method DeepCAD~\cite{DeepCAD}, our approach shows clear advantages across all evaluated metrics. In the qualitative comparison with DeepCAD~\cite{DeepCAD}, as illustrated in Fig.~\ref{fig:recon_exp2_gen}, our method effectively generates the correct command sequence conditioned on the corresponding point cloud in the majority of cases. In contrast, DeepCAD struggles in several instances, particularly in the generation quality regarding fine details.

\begin{table}[]
\setlength{\tabcolsep}{4pt} %

\centering
\resizebox{\linewidth}{!}{
\begin{tabular}{c|ccc}
\toprule
\multicolumn{1}{c|}{Methods}      & Chamfer($\times 100$)$\downarrow$        & F-score($\times 100$)$\uparrow$  & Normal C($\times 100$)$\uparrow$
\\ \hline

 InstantMesh~\cite{xu2024instantmesh}  & 5.38   & 61.81   & 45.53   
\\

 Ours(Image) & \textbf{3.77} & \textbf{76.70} & \textbf{59.62} 
\\

\bottomrule
\end{tabular}
}

\caption{The quantitative results on image-based reconstruction tasks. We observe that our method outperforms reconstruction metrics.
}
\label{table:image_recon}
\end{table}

\begin{figure}[h]
\centering
\includegraphics[width=0.45\textwidth]{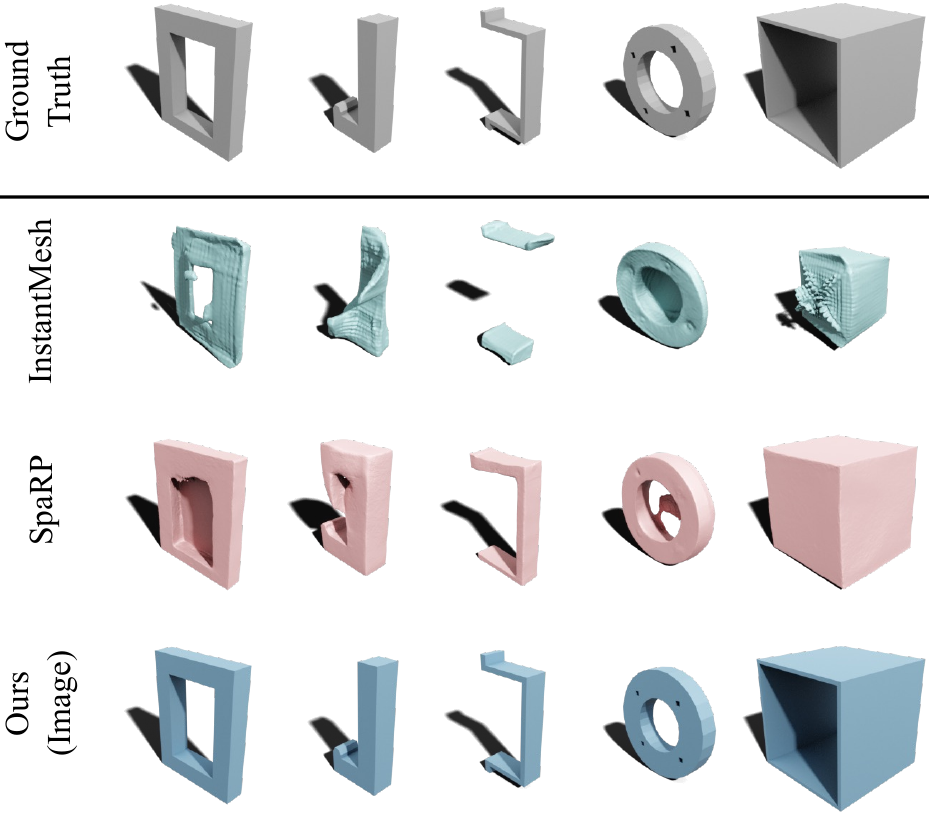}
\caption{The qualitative comparison with image-to-mesh baselines. InstantMesh\cite{xu2024instantmesh} struggles to reconstruct the model's shape accurately. While SpaRP\cite{xu2024sparp} manages to capture the rough shape, it falls short of producing a smooth and axis-aligned CAD model.}
\label{fig:image_recon}
\end{figure}

\begin{figure*}[h]
\centering
\includegraphics[width=0.95\textwidth]{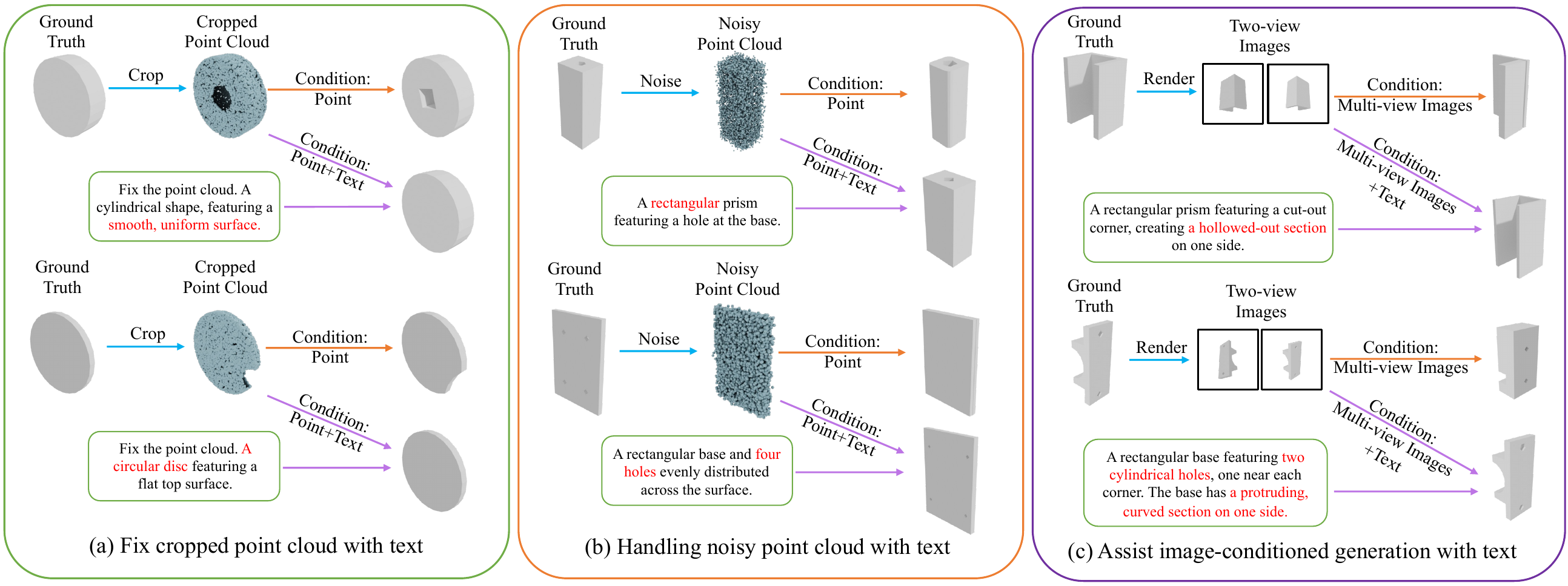}
\caption{We present three different applications of our multimodal model. (a) When generating CAD models conditioned solely on the cropped point cloud, our multimodal model will be influenced by the missing spatial information of the cropped data. After additionally prompted with the description of the complete CAD model, our multimodal model is able to fix the cropping. (b) When generating CAD models conditioned solely on the noisy point cloud, our multimodal model will be influenced by the missing spatial details. After additionally prompted with the description of the original CAD model, our multimodal model is able to retain the original feature. (c) When generating CAD models conditioned solely on the two-view images, the complete CAD models are not fully observed. Our multimodal model may fail in some cases. After additionally prompted with the description of the full CAD model, our multimodal model is able to fill the unobserved geometry.}
\label{fig:recon_text_fix}
\end{figure*}

\subsubsection{Image Conditioned CAD Generation}
Similar to the training setting of point-based CAD reconstruction, we utilize images exclusively as inputs for training. To ensure a fair comparison, we configure the baseline inference to also process two-view images, aligning with our experimental setup. Given that SpaRP~\cite{xu2024sparp} has not been open-sourced and provides only a web demo for inference, our comparisons with it are qualitative only.

We quantitatively compare our method with InstantMesh~\cite{xu2024instantmesh} and show the result in Tab.~\ref{table:image_recon}. Note that InstantMesh~\cite{xu2024instantmesh} benefits from an iso-surface extraction module~\cite{shen2023flexicubes}, which inherently produces meshes with excellent connectivity. This leads to notably low DangEL and SIR values in its extraction process. Furthermore, the use of Signed Distance Functions (SDF) ensures watertight geometries, resulting in exceptionally low FluxEE values. Given these methodological differences and distinct focus areas, our comparison just focuses on the reconstruction metrics. We can observe that our method performs better on reconstruction metrics.

For the qualitative comparison, we evaluate both SpaRP\cite{xu2024sparp} and InstantMesh\cite{xu2024instantmesh}. From Fig.~\ref{fig:image_recon}, we observe that InstantMesh struggles with accurately reconstructing the model’s shape, while SpaRP captures the general structure but fails to deliver a smooth, axis-aligned result. In contrast, our method successfully reconstructs a smooth and precise CAD representation.

\subsubsection{Text Conditioned CAD Generation}
Since there are currently no established metrics specifically designed to evaluate the generation of CAD models conditioned on textual descriptions, we compare our method with the open-source method Michelangelo~\cite{zhao2023michelangelo} and the closed-source website Tripo~\cite{tripo3d_ai} by conducting a user study. We invite 16 participants to evaluate 10 pairs of text descriptions and their corresponding generated 3D models. The participants are asked to rate the models based on two criteria: the alignment between the models and the conditioned textual descriptions and the overall quality of the generated models. Each data pair should be scored on a scale of 1 to 5, where 1 represents the lowest quality and 5 represents the highest. Tab.~\ref{table:text_user_study} presents the average scores. Our method achieves the highest scores in terms of text alignment and is comparable to that of Tripo in terms of model quality.

\begin{table}[]
\setlength{\tabcolsep}{2pt} %

\centering
\resizebox{0.7\linewidth}{!}{
\begin{tabular}{c|cc}
\toprule
\multicolumn{1}{c|}{Methods}      &     Text Alignment    & Model Quality    
\\ \hline

Michelangelo~\cite{zhao2023michelangelo} & 1.16 & 2.04
\\

Tripo~\cite{tripo3d_ai} & 3.30 & \textbf{4.58}
\\

Ours(Text) & \textbf{4.16} & 4.45
\\
\bottomrule
\end{tabular}
}
\caption{The user study results on text-conditioned generation tasks. We evaluate the generated models based on two criteria: text alignment and model quality. A higher score indicates better performance, with both criteria rated on a 5-point scale. }
\label{table:text_user_study}
\end{table}

\begin{figure}[h]
\centering
\includegraphics[width=0.45\textwidth]{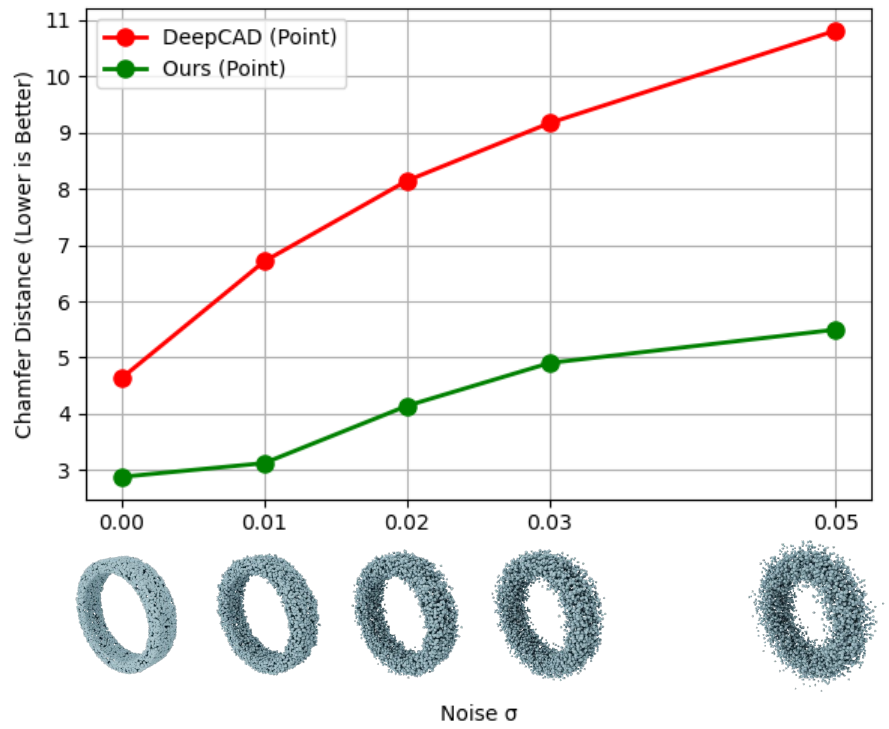}
\caption{The Chamfer Distance under varying noise levels. While both baseline~\cite{DeepCAD} and our quality degrade with noises, our approach demonstrates stronger robustness in handling noisy point cloud data.}
\label{fig:noise_chart}
\end{figure}

\subsubsection{Multimodal-Input-Conditioned CAD Generation}

In Fig.~\ref{fig:recon_text_fix}, we present three scenarios to demonstrate our multimodal model's adaptability in CAD generation across different input conditions.

\noindent \textbf{(a) Cropped Point Cloud:} When using only a cropped point cloud, our model will be influenced by the partial lack of spatial information. Supplementing this input with a complete CAD model description enables the model to compensate, reconstructing missing areas effectively.

\noindent \textbf{(b) Noisy Point Cloud:} Noisy point clouds reduce detail accuracy. By including a descriptive prompt of the original CAD model, the model produces a more accurate output.

\noindent \textbf{(c) Two-View Images:} With two-view images, incomplete viewpoints may lead to missing geometry. Adding a full model description helps the model fill in unobserved sections, achieving a more complete CAD generation.

These cases highlight our model's strength in leveraging multimodal inputs to address challenges from partial or noisy data, enhancing CAD generation fidelity and completeness.

\subsection{Robustness Evaluation}

To further assess the robustness of our approach, we conduct experiments on two challenging tasks: point cloud data with added noise and point cloud data with random point elimination. For each task, we randomly select 1,000 cases from the test set. The noisy point cloud tests examine how well our model can generate CAD models under varying levels of perturbation in the input data, while the partial point cloud tests evaluate the model's ability to reconstruct accurate shapes with less data.

\begin{figure}[h]
\centering
\includegraphics[width=0.45\textwidth]{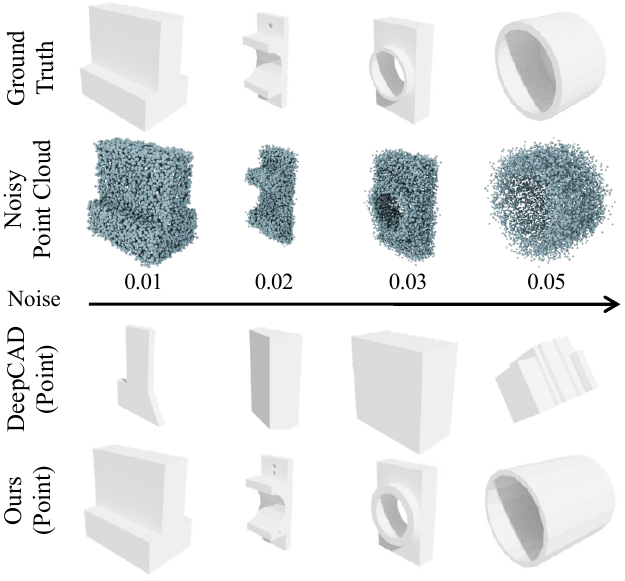}
\caption{The qualitative evaluation with different levels of noises added to the point clouds. While our method shows some errors due to the noises, the overall shape structure remains well-preserved. In contrast, DeepCAD~\cite{DeepCAD} is more significantly affected by the noises.}
\label{fig:noise_exp}
\end{figure}

\begin{figure}[h]
\centering
\includegraphics[width=0.45\textwidth]{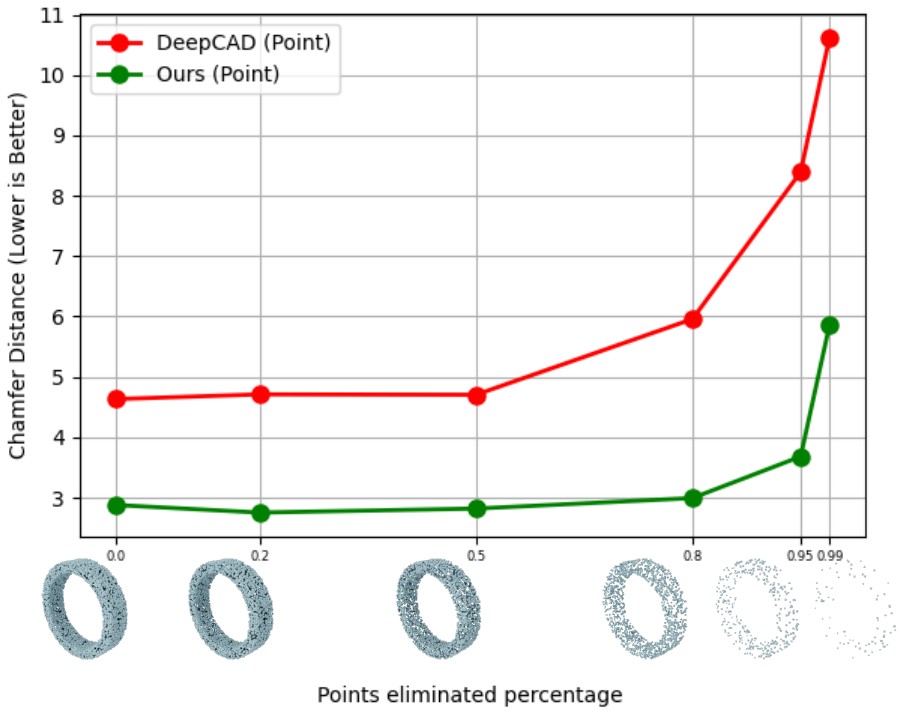}
\caption{The Chamfer Distance under different point elimination percentages. Our method demonstrates superior robustness across all point reduction levels. Even if 95\% points are removed, our method can still robustly recover the correct CAD models.
}
\label{fig:part_chart}
\end{figure}

\subsubsection{Performance on Noisy Point Cloud Inputs}

To simulate noisy data, we sample noises from a normal distribution, with zero mean and standard deviation of $\sigma$, as an offset of the points in the three positional dimensions.

We use Chamfer Distance as the primary metric for evaluating reconstruction performance and present the results in Fig.\ref{fig:noise_chart}, comparing our model with DeepCAD~\cite{DeepCAD} under varying noise levels. As observed, although both methods experience a decline in performance as noise levels increase, our model demonstrates a slower degradation, indicating stronger robustness when handling noisy point cloud data. A similar trend is visible in Fig.\ref{fig:noise_exp}, where we compare the qualitative robustness of our approach against DeepCAD. While noise introduces some errors in the fine details of our model's reconstruction, the overall structural information remains largely consistent with the ground truth. In contrast, DeepCAD is more significantly affected by the presence of noise, resulting in poorer reconstruction performance. Comprehensive quantitative results for all metrics across all tested noise levels are provided in the supplementary material.

\begin{figure}[h]
\centering
\includegraphics[width=0.45\textwidth]{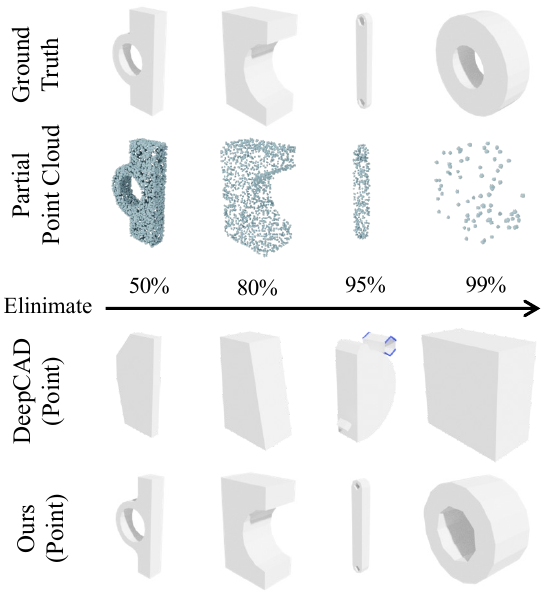}
\caption{The qualitative evaluation with different percentages of points eliminated. Our method reconstructs the overall shape well over different eliminated percentages, while DeepCAD~\cite{DeepCAD} is more adversely affected by the elimination.}
\label{fig:part_exp}
\end{figure}

\subsubsection{Performance on Partial Point Cloud Inputs}

To evaluate the model's performance when only partial point cloud data is provided, we progressively eliminate different percentages of points randomly from the original point cloud.

In Fig.~\ref{fig:part_chart}, we compare our method with DeepCAD~\cite{DeepCAD} under various levels of point cloud reduction in terms of chamfer distance. Our method consistently outperforms DeepCAD across all reduction levels. Surprisingly, even when 95\% of the point cloud data is removed, our model still maintains strong reconstruction performance, outperforming DeepCAD's results on complete point clouds.

\begin{table*}[]
\setlength{\tabcolsep}{4pt} %

\centering
\resizebox{\linewidth}{!}{
\begin{tabular}{c|ccccccc}
\toprule
\multicolumn{1}{c|}{Methods}      & Chamfer($\times 100$)$\downarrow$        & F-score($\times 100$)$\uparrow$  & Normal C($\times 100$)$\uparrow$ & SegE$\downarrow$ & DangEL$\downarrow$ & SIR(\%)$\downarrow$   &   FluxEE($\times 100$)$\downarrow$        \\ \hline

 DeepCAD~\cite{DeepCAD}(Point)  & 7.61    & 52.77     & 51.96   & 13.77  & 2.13 & 9.12  &  0.276  \\

 Ours(Point)    & \textbf{3.39} & \textbf{79.27} & \textbf{66.78} & \textbf{2.05} &  \textbf{0.63} &  \textbf{1.68}  & \textbf{0.194}  \\

\bottomrule
\end{tabular}
}

\caption{Quantitative generalization test on Fusion360 reconstruction dataset~\cite{fusion360recon}. Our method outperforms DeepCAD~\cite{DeepCAD} across all reconstruction, topology, and enclosure metrics on the unseen data. Neither our method nor DeepCAD is trained using Fusion360 data.
}
\label{table:out_of_distribution_exp}
\end{table*}

In Fig.~\ref{fig:part_exp}, we visualize the qualitative robustness evaluation with varying percentages of points eliminated from the original point clouds. As expected, the reduction in points impacts the expressiveness of the shape, but remarkably, even under extreme conditions where 99\% of the points are removed, our model still accurately reconstructs the overall layout and structure of the CAD model. Although there are slight deviations in fine details and dimensions, the core geometry is preserved. In contrast, DeepCAD~\cite{DeepCAD} is significantly more affected by the reduction, leading to much poorer reconstruction results. This highlights the robustness of our method in handling sparse point clouds. Comprehensive quantitative results for all metrics across all tested elimination percentages are provided in the supplementary material.

\subsection{Generalization Assessment}

\begin{figure}[h]
\centering
\includegraphics[width=0.45\textwidth]{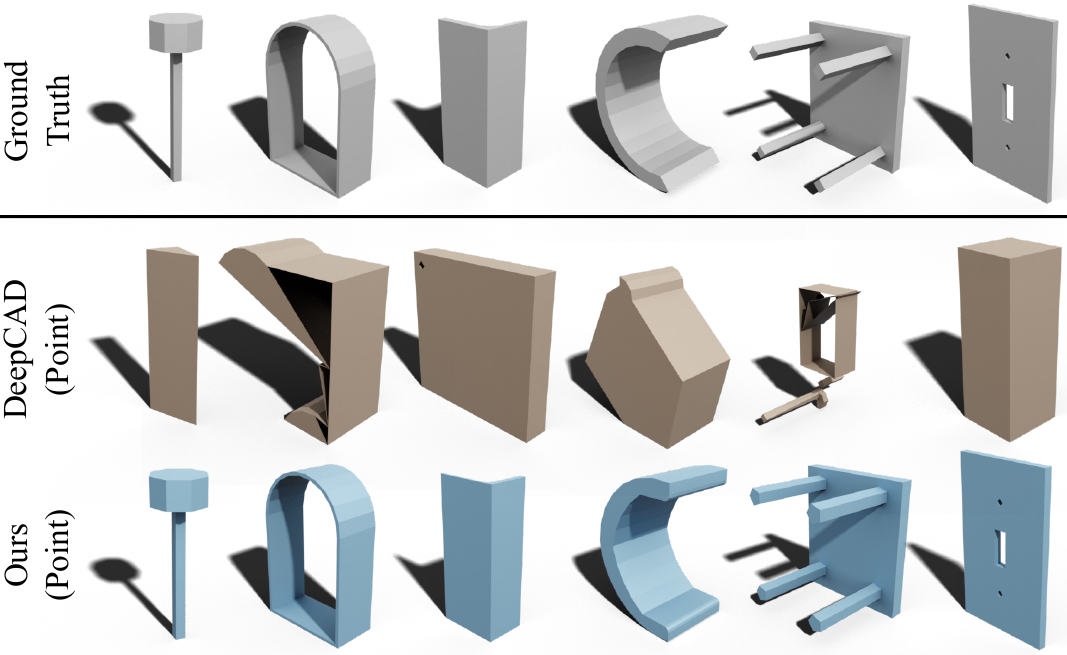}
\caption{We present qualitative comparisons for the generalization test on Fusion360~\cite{fusion360recon}, demonstrating that our method achieves better reconstruction of models than DeepCAD~\cite{DeepCAD}. Notably, neither our method nor DeepCAD is trained using Fusion360 data.}
\label{fig:out_of_distribution_exp}
\end{figure}

To validate the generalization performance of our method on unseen data, we conduct tests using the Fusion360 reconstruction dataset~\cite{fusion360recon}. We randomly sample 1,512 models from the dataset to create a test set, utilizing point cloud data as input for evaluation. The results of our tests are presented quantitatively and qualitatively in Tab.~\ref{table:out_of_distribution_exp} and Fig~\ref{fig:out_of_distribution_exp}, respectively. Our method consistently outperforms the baseline across all evaluation metrics, demonstrating the effective generalization capabilities on unseen data.

\vspace{-5pt}

\begin{figure*}[h]
\centering
\includegraphics[width=0.95\textwidth]{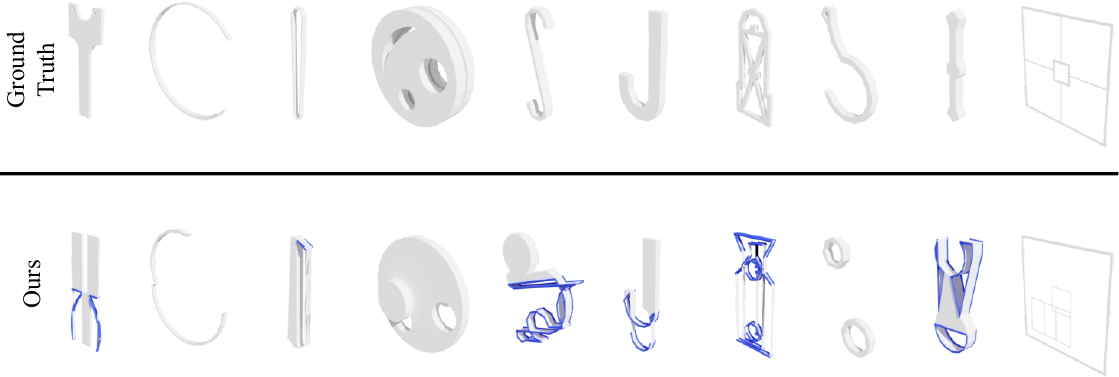}
\caption{We present the failure case of our method on some challenging cases, such as thin structures and complicated details. Dangling edges exist in these failure results.}
\label{fig:recon_recon_failure}
\end{figure*}

\begin{table*}[]
\setlength{\tabcolsep}{4pt} %

\centering

{\small (a) Image Reconstruction Comparison} 
\resizebox{\linewidth}{!}{
\begin{tabular}{c|ccccccc}
\toprule
\multicolumn{1}{c|}{Methods}      & Chamfer($\times 100$)$\downarrow$        & F-score($\times 100$)$\uparrow$  & Normal C($\times 100$)$\uparrow$ & SegE$\downarrow$ & DangEL$\downarrow$ & SIR(\%)$\downarrow$   &   FluxEE($\times 100$)$\downarrow$        \\ \hline

 Ours(Image)  & 3.77   & 76.70   & 59.62   & 1.97  & 0.79 & 2.07  &  0.063  \\
 Ours(Multimodal)  & \textbf{3.22} & \textbf{80.82} & \textbf{62.07} & \textbf{1.56} &  \textbf{0.51} &  \textbf{1.36}  & \textbf{0.050}  \\
\bottomrule
\end{tabular}
}

\vspace{1mm}
{\small (b) Point Reconstruction Comparison} 

\resizebox{\linewidth}{!}{
\begin{tabular}{c|ccccccc}
\toprule
\multicolumn{1}{c|}{Methods}      & Chamfer($\times 100$)$\downarrow$        & F-score($\times 100$)$\uparrow$  & Normal C($\times 100$)$\uparrow$ & SegE$\downarrow$ & DangEL$\downarrow$ & SIR(\%)$\downarrow$   &   FluxEE($\times 100$)$\downarrow$        \\ \hline

 Ours(Point)    & \textbf{1.85} & \textbf{90.88} & \textbf{79.71} & 1.66 &  \textbf{0.46} &  \textbf{1.31}  & 0.044  \\
  Our(Multimodal)  & 2.63   & 85.17   & 73.64   & \textbf{1.53}  & 0.47 & 1.32  &  \textbf{0.035}  \\

\bottomrule
\end{tabular}
}

\caption{The quantitative study on training with multimodal data. (a) Our multimodal model outperforms the image-only model across all reconstruction, topology, and enclosure metrics. (b) Our multimodal model is comparable with the point-only model in terms of topology and enclosure metrics. However, the point reconstruction performance of our multimodal model is slightly weaker than the point-only model.}
\label{table:multimodal}
\end{table*}

\subsection{Influence of Multimodal Data Training}
Additionally, we investigate the impact of training with multimodal data compared to using single-modal training data on the generated CAD models. As shown in Tab.~\ref{table:multimodal}, we compare the performance of models trained on multimodal datasets against those trained exclusively on image data and those trained solely on point cloud data. To ensure fairness in our testing, we utilize corresponding single-modal data for the evaluation phase.

We can observe from Tab.~\ref{table:multimodal}(a) that when testing on datasets where only images are used as input conditions, models trained on multimodal data outperform those trained solely on image data. We hypothesize that this improvement is due to the additional complementary information provided by other modalities, such as point clouds, which offer more detailed insights into the structure and geometry of the CAD models. This enriched information helps the model to generate higher-quality CAD models. 

Tab.~\ref{table:multimodal}(b) shows that when conditioned solely on point data, our multimodal data trained model achieves comparable performance with the model trained exclusively on point cloud data from the topology and enclosure perspective. However, the point reconstruction accuracy of the only point data trained model surpasses that of the multimodal data trained model. A possible reason for this observation is that in unimodal training using only point cloud data, the model can fully focus on optimizing the point cloud representation. And the point cloud data possesses sufficient CAD model's detailed information, resulting in higher point reconstruction accuracy. In contrast, models trained on multimodal data must integrate representations from various modalities and balance the optimization across them. The introduction of other modalities, particularly textual descriptions, which are inherently more coarse and less precise, may introduce noise into the training process. This noise can negatively impact the accuracy of point reconstruction.

\section{Limitations}
Despite the promising and robust performance of CAD-MLLM, several limitations exist. First, while InternVL2-26B is a commendable open-source multimodal large model, it is particularly sensitive to perspective distortions in multi-view images, especially when dealing with complex shapes, which can adversely affect the generation of textual descriptions. Additionally, current text descriptions often fail to accurately capture the precise geometry of complex shapes, primarily due to a lack of specific CAD dimension information. This results in relative descriptions of attributes such as edge lengths or apertures, rather than absolute size measurements. Consequently, the generated CAD models may exhibit similar shapes but differ significantly in size. This limitation may be addressed by leveraging other work, like  ~\cite{khan2024automaticfeaturerecognitiondimensional}, to extract CAD dimension attributes from the training data, thereby enabling more precise dimension information to be incorporated during model training. Some failure cases are illustrated in Fig.\ref{fig:recon_recon_failure}.

\section{Conclusions}
In this work, we propose CAD-MLLM, a MLLM-assisted framework designed to generate parametric CAD models based on textual descriptions, multi-view images, point clouds, or any combination of these inputs, thus facilitating ease of use for non-expert users. To tackle this challenging task, we factor it into two sub-problems. First, we explore a vectorized representation of CAD command sequences to enhance LLM understanding, aligning the feature spaces of multi-view images and point clouds within the LLM's framework. Additionally, to address the gaps in existing datasets regarding multimodality information and to empower LLM capabilities, we propose a new dataset, Omni-CAD. We evaluate our method on this dataset, and beyond traditional reconstruction quality metrics, we introduce four novel evaluation criteria that focus on topology quality and surface enclosure extent. Extensive experimental results demonstrate that our approach outperforms the previous generation methods while exhibiting greater robustness.

\section{Acknowledgement}

We sincerely thank Rundi Wu and Xiang Xu for clarifying certain questions during our code implementation process.

Additionally, we thank Ruihan Yu and Shangzhe Li for inspiring discussions on field theory in physics, which led to sparking inspiration for part of this paper.

\ifCLASSOPTIONcaptionsoff
  \newpage
\fi

\bibliographystyle{IEEEtran}
\bibliography{ref}

\clearpage


\end{document}


\title{Supplementary material of CAD-MLLM:\\ Unifying Multimodality-Conditioned\\ CAD Generation With MLLM}

\author{
        Jingwei~Xu*,
        Chenyu~Wang*,
        Zibo~Zhao,
        Wen~Liu,
        Yi~Ma,
        Shenghua~Gao
\IEEEcompsocitemizethanks{
\IEEEcompsocthanksitem Jingwei Xu and Chenyu Wang contributed equally to this work;
\IEEEcompsocthanksitem  Corresponding Author: Shenghua Gao; \protect\\ 
E-mail: gaosh@hku.hk
\IEEEcompsocthanksitem Jingwei Xu and Zibo Zhao are with the School of Information Science and Technology, ShanghaiTech University, Shanghai 201210, China. Email: xujw2023@shanghaitech.edu.cn, zhaozb@shanghaitech.edu.cn
\IEEEcompsocthanksitem Chenyu Wang is with Transcengram. Email: wangchy@transcengram.com
\IEEEcompsocthanksitem Wen Liu is with DeepSeek AI. Email: liuwen@deepseek.com
\IEEEcompsocthanksitem Yi Ma, and Shenghua Gao are with the University of Hong Kong, Hong Kong SAR,
China. E-mail: mayi@hku.hk, gaosh@hku.hk

}%
}

\maketitle

\section{Quantitative Results of Robustness Tests}

We provide the complete quantitative evaluation results of both \textbf{Noisy Point Cloud Test} and \textbf{Partial Point Cloud Test} in Tab.~\ref{table:point_robustness_1} and Tab.~\ref{table:point_robustness_2}. Our method outperforms DeepCAD~[1] across all metrics at various kinds and levels of data flaws, which indicates the better robustness of our method.

\FloatBarrier

\begin{table}[H]
\setlength{\tabcolsep}{4pt} %

\centering

{\small (A) Clean data} 

\resizebox{\linewidth}{!}{
\begin{tabular}{c|ccccccc}
\toprule

 \multicolumn{1}{c|}{\multirow{2}{*}{Methods}}   & Chamfer   & F-score & Normal C & \multirow{2}{*}{SegE$\downarrow$} & \multirow{2}{*}{DangEL$\downarrow$} & SIR  &   FluxEE   \\
& ($\times 100$)$\downarrow$ & ($\times 100$)$\uparrow$ &  ($\times 100$)$\uparrow$  & & & (\%)$\downarrow$ &  ($\times 100$)$\downarrow$\\
\hline

 DeepCAD~[1](Point)  & 4.63  &  71.47  &  64.47  & 9.47 & 1.32  &  6.35  &  0.375  \\
 Ours(Point)    & \textbf{2.88} & \textbf{83.10} & \textbf{72.66} & \textbf{2.22} &  \textbf{0.64} &  \textbf{2.02}  & \textbf{0.066}  \\

\bottomrule
\end{tabular}
}

\vspace{4mm}
{\small (B$_1$) Noisy data with $\sigma_2=0.01$} 
\resizebox{\linewidth}{!}{
\begin{tabular}{c|ccccccc}
\toprule

 \multicolumn{1}{c|}{\multirow{2}{*}{Methods}}   & Chamfer   & F-score & Normal C & \multirow{2}{*}{SegE$\downarrow$} & \multirow{2}{*}{DangEL$\downarrow$} & SIR  &   FluxEE   \\
& ($\times 100$)$\downarrow$ & ($\times 100$)$\uparrow$ &  ($\times 100$)$\uparrow$  & & & (\%)$\downarrow$ &  ($\times 100$)$\downarrow$\\
\hline

 DeepCAD~[1](Point)  & 6.71   & 55.97   & 53.34   & 9.27  & 1.38 & 8.97  &  0.227  \\
 Ours(Point)    & \textbf{3.12} & \textbf{82.05} & \textbf{71.11} & \textbf{2.21} &  \textbf{0.70} &  \textbf{1.85}  & \textbf{0.025}  \\

\bottomrule
\end{tabular}
}

\vspace{1mm}
{\small (B$_2$) Noisy data with $\sigma_2=0.02$} 

\resizebox{\linewidth}{!}{
\begin{tabular}{c|ccccccc}
\toprule

 \multicolumn{1}{c|}{\multirow{2}{*}{Methods}}   & Chamfer   & F-score & Normal C & \multirow{2}{*}{SegE$\downarrow$} & \multirow{2}{*}{DangEL$\downarrow$} & SIR  &   FluxEE   \\
& ($\times 100$)$\downarrow$ & ($\times 100$)$\uparrow$ &  ($\times 100$)$\uparrow$  & & & (\%)$\downarrow$ &  ($\times 100$)$\downarrow$\\
\hline

 DeepCAD~[1](Point)  & 8.15   & 46.67   & 49.64   & 16.99  & 1.94 & 7.63  &  0.511  \\
 Ours(Point)    & \textbf{4.14} & \textbf{74.39} & \textbf{65.66} & \textbf{2.31} &  \textbf{0.51} &  \textbf{1.82}  & \textbf{0.049}  \\

\bottomrule
\end{tabular}
}

\vspace{1mm}
{\small (B$_3$) Noisy data with $\sigma_2=0.03$} 

\resizebox{\linewidth}{!}{
\begin{tabular}{c|ccccccc}
\toprule

 \multicolumn{1}{c|}{\multirow{2}{*}{Methods}}   & Chamfer   & F-score & Normal C & \multirow{2}{*}{SegE$\downarrow$} & \multirow{2}{*}{DangEL$\downarrow$} & SIR  &   FluxEE   \\
& ($\times 100$)$\downarrow$ & ($\times 100$)$\uparrow$ &  ($\times 100$)$\uparrow$  & & & (\%)$\downarrow$ &  ($\times 100$)$\downarrow$\\
\hline

 DeepCAD~[1](Point)  & 9.17   & 40.84   & 45.83   & 16.75  & 2.01 & 10.10  &  0.363  \\
 Ours(Point)    & \textbf{4.91} & \textbf{68.99} & \textbf{61.51} & \textbf{3.96} &  \textbf{0.81} &  \textbf{2.86}  & \textbf{0.283}  \\

\bottomrule
\end{tabular}
}

\vspace{1mm}
{\small (B$_4$) Noisy data with $\sigma_2=0.05$} 

\resizebox{\linewidth}{!}{
\begin{tabular}{c|ccccccc}
\toprule

 \multicolumn{1}{c|}{\multirow{2}{*}{Methods}}   & Chamfer   & F-score & Normal C & \multirow{2}{*}{SegE$\downarrow$} & \multirow{2}{*}{DangEL$\downarrow$} & SIR  &   FluxEE   \\
& ($\times 100$)$\downarrow$ & ($\times 100$)$\uparrow$ &  ($\times 100$)$\uparrow$  & & & (\%)$\downarrow$ &  ($\times 100$)$\downarrow$\\
\hline

 DeepCAD~[1](Point)  & 10.82   & 32.69   & 44.02   & 14.70  & 2.44 &  13.54 & 1.230  \\
 Ours(Point)    & \textbf{5.50} & \textbf{63.76} & \textbf{57.03} & \textbf{3.88} &  \textbf{0.99} &  \textbf{3.51}  & \textbf{0.199}  \\

\bottomrule
\end{tabular}
}

\caption{The quantitative experiment of the robustness tests with noisy data. We observe that over different noise levels, our method demonstrates greater robustness than DeepCAD~[1] across all metrics.}
\label{table:point_robustness_1}
\end{table}

\FloatBarrier

\section{Multimodal Conditioned Input Dataset Visualization}
\begin{figure*}[ht]
\centering
\includegraphics[width=0.95\textwidth]{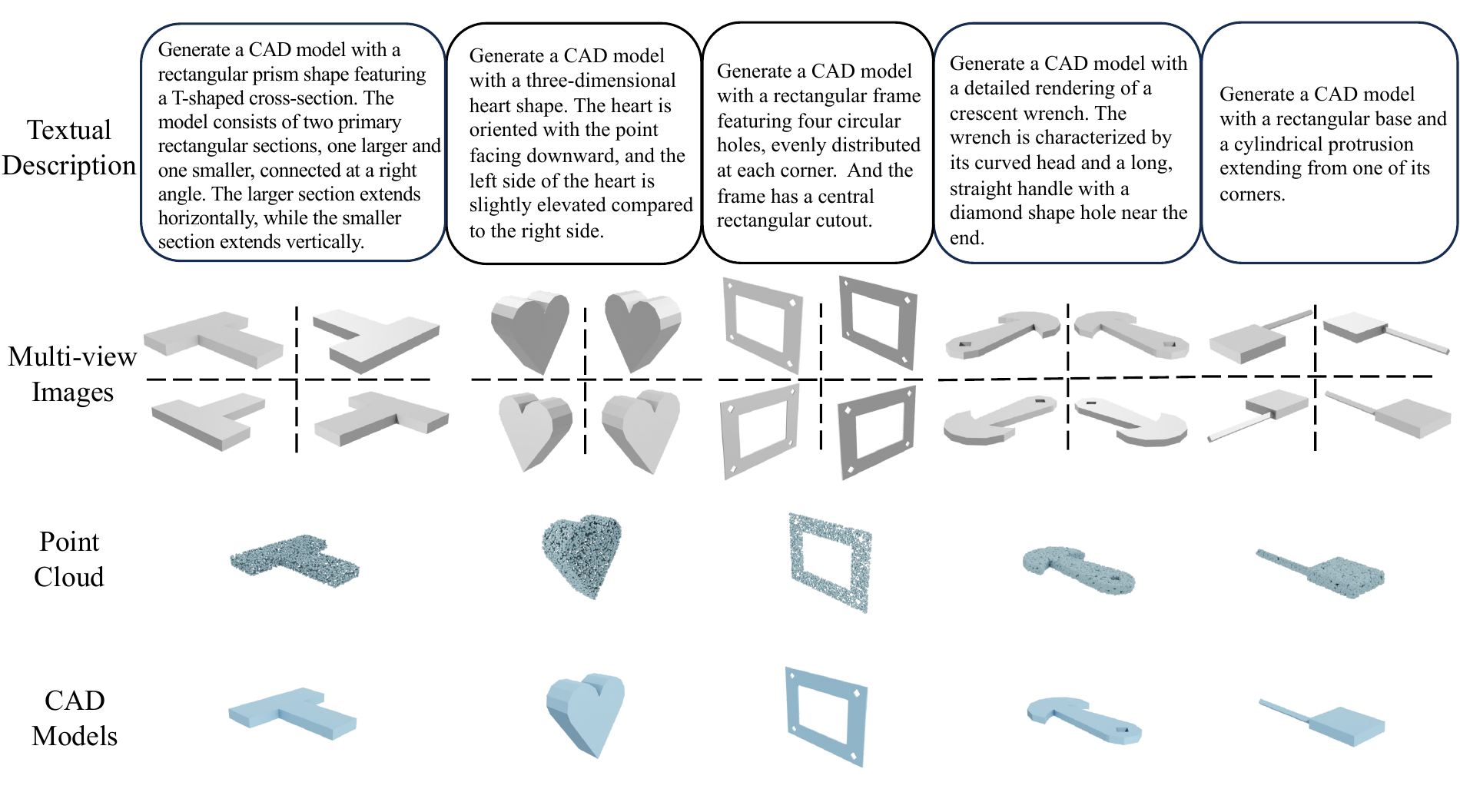}
\caption{\textbf{Dataset sample visualization.} We sample five cases from our proposed Omni-CAD dataset to illustrate the multimodal conditioned data and the corresponding ground truth CAD models. In the real dataset, each CAD model includes images of eight views; here, we randomly select four views for demonstration purposes.}
\label{fig:dataset}
\end{figure*}

We illustrate 5 pairs of data samples in Fig.~\ref{fig:dataset}. As mentioned in the main text, we construct corresponding multimodal data for each CAD model, including textual descriptions, images rendered from 8 fixed angles, and point cloud data. Here, we randomly selected images from 4 of these 8 angles for visualization purposes.

\begin{table}[H]
\setlength{\tabcolsep}{4pt} %

\centering

{\small (A) Clean data} 

\resizebox{\linewidth}{!}{
\begin{tabular}{c|ccccccc}
\toprule

 \multicolumn{1}{c|}{\multirow{2}{*}{Methods}}   & Chamfer   & F-score & Normal C & \multirow{2}{*}{SegE$\downarrow$} & \multirow{2}{*}{DangEL$\downarrow$} & SIR  &   FluxEE   \\
& ($\times 100$)$\downarrow$ & ($\times 100$)$\uparrow$ &  ($\times 100$)$\uparrow$  & & & (\%)$\downarrow$ &  ($\times 100$)$\downarrow$\\
\hline

 DeepCAD~[1](Point)  & 4.63  &  71.47  &  64.47  & 9.47 & 1.32  &  6.35  &  0.375  \\
 Ours(Point)    & \textbf{2.88} & \textbf{83.10} & \textbf{72.66} & \textbf{2.22} &  \textbf{0.64} &  \textbf{2.02}  & \textbf{0.066}  \\

\bottomrule
\end{tabular}
}

\vspace{4mm}
{\small (C$_1$) Eliminate $20\%$ points} 

\resizebox{\linewidth}{!}{
\begin{tabular}{c|ccccccc}
\toprule

 \multicolumn{1}{c|}{\multirow{2}{*}{Methods}}   & Chamfer   & F-score & Normal C & \multirow{2}{*}{SegE$\downarrow$} & \multirow{2}{*}{DangEL$\downarrow$} & SIR  &   FluxEE   \\
& ($\times 100$)$\downarrow$ & ($\times 100$)$\uparrow$ &  ($\times 100$)$\uparrow$  & & & (\%)$\downarrow$ &  ($\times 100$)$\downarrow$\\
\hline

 DeepCAD~[1](Point)  &  4.71  &  71.47  & 64.63   &  7.64 & 1.34 & 6.03 &  0.281 \\
 Ours(Point)    & \textbf{2.75} & \textbf{84.79} & \textbf{73.44} & \textbf{2.17} &  \textbf{0.36} &  \textbf{1.89}  & \textbf{0.138}  \\

\bottomrule
\end{tabular}
}

\vspace{1mm}
{\small (C$_2$) Eliminate $50\%$ points} 

\resizebox{\linewidth}{!}{
\begin{tabular}{c|ccccccc}
\toprule

 \multicolumn{1}{c|}{\multirow{2}{*}{Methods}}   & Chamfer   & F-score & Normal C & \multirow{2}{*}{SegE$\downarrow$} & \multirow{2}{*}{DangEL$\downarrow$} & SIR  &   FluxEE   \\
& ($\times 100$)$\downarrow$ & ($\times 100$)$\uparrow$ &  ($\times 100$)$\uparrow$  & & & (\%)$\downarrow$ &  ($\times 100$)$\downarrow$\\
\hline

 DeepCAD~[1](Point)  & 4.70   &  71.40  &  64.19  & 8.88  & 1.41 & 5.33 & 0.138  \\
 Ours(Point)    & \textbf{2.82} & \textbf{83.37} & \textbf{72.69} & \textbf{2.14} &  \textbf{0.45} &  \textbf{1.67}  & \textbf{0.025}  \\

\bottomrule
\end{tabular}
}

\vspace{1mm}
{\small (C$_3$) Eliminate $80\%$ points} 

\resizebox{\linewidth}{!}{
\begin{tabular}{c|ccccccc}
\toprule

 \multicolumn{1}{c|}{\multirow{2}{*}{Methods}}   & Chamfer   & F-score & Normal C & \multirow{2}{*}{SegE$\downarrow$} & \multirow{2}{*}{DangEL$\downarrow$} & SIR  &   FluxEE   \\
& ($\times 100$)$\downarrow$ & ($\times 100$)$\uparrow$ &  ($\times 100$)$\uparrow$  & & & (\%)$\downarrow$ &  ($\times 100$)$\downarrow$\\
\hline

 DeepCAD~[1](Point)  &  5.96  & 62.32   &  58.40  & 12.51  &  1.44 & 7.54 &  0.462 \\
 Ours(Point)    & \textbf{2.99} & \textbf{82.82} & \textbf{71.90} & \textbf{2.43} &  \textbf{0.66} &  \textbf{1.74}  & \textbf{0.086}  \\

\bottomrule
\end{tabular}
}

\vspace{1mm}
{\small (C$_4$) Eliminate $95\%$ points} 

\resizebox{\linewidth}{!}{
\begin{tabular}{c|ccccccc}
\toprule

 \multicolumn{1}{c|}{\multirow{2}{*}{Methods}}   & Chamfer   & F-score & Normal C & \multirow{2}{*}{SegE$\downarrow$} & \multirow{2}{*}{DangEL$\downarrow$} & SIR  &   FluxEE   \\
& ($\times 100$)$\downarrow$ & ($\times 100$)$\uparrow$ &  ($\times 100$)$\uparrow$  & & & (\%)$\downarrow$ &  ($\times 100$)$\downarrow$\\
\hline

 DeepCAD~[1](Point)  &  8.39  &  44.86  &  47.70  & 18.28  & 1.73  & 7.75 & 0.560  \\
 Ours(Point)    & \textbf{3.68} & \textbf{76.73} & \textbf{65.43} & \textbf{2.44} &  \textbf{0.71} &  \textbf{1.92}  & \textbf{0.040}  \\

\bottomrule
\end{tabular}
}

\vspace{1mm}
{\small (C$_5$) Eliminate $99\%$ points} 

\resizebox{\linewidth}{!}{
\begin{tabular}{c|ccccccc}
\toprule

 \multicolumn{1}{c|}{\multirow{2}{*}{Methods}}   & Chamfer   & F-score & Normal C & \multirow{2}{*}{SegE$\downarrow$} & \multirow{2}{*}{DangEL$\downarrow$} & SIR  &   FluxEE   \\
& ($\times 100$)$\downarrow$ & ($\times 100$)$\uparrow$ &  ($\times 100$)$\uparrow$  & & & (\%)$\downarrow$ &  ($\times 100$)$\downarrow$\\
\hline

 DeepCAD~[1](Point)  &  10.62  &  34.02  &  44.14  &  7.71 & 1.32  &  7.54  &  0.323 \\
 Ours(Point)    & \textbf{5.86} & \textbf{60.08} & \textbf{54.07} & \textbf{2.83} &  \textbf{0.26} &  \textbf{1.60}  & \textbf{0.005}  \\

\bottomrule
\end{tabular}
}

\caption{The quantitative experiment of the robustness tests with noisy data. We observe that over different percentages of eliminated point clouds, our method demonstrates greater robustness than DeepCAD~[1] across all metrics.}
\label{table:point_robustness_2}
\end{table}

\bibliographystyle{IEEEtran}